\documentclass[final,5p,times,twocolumn]{elsarticle}

\usepackage{amssymb}
\usepackage{amsmath}
\usepackage{multirow}
\usepackage{caption}
\usepackage{subcaption}
\usepackage{tabularx}
\usepackage{natbib}
\usepackage{hyperref}
\usepackage{color}
\usepackage{pifont}

\journal{Knowledge-Based Systems}

\begin{document}
\begin{frontmatter}

\title{Multimodality in Meta-Learning: A Comprehensive Survey}

\cortext[mycorrespondingauthor]{Corresponding author}
\author[inst1,inst2]{Yao Ma}
\ead{Y.MA-11@student.tudelft.nl}
\author[inst1]{Shilin Zhao}
\ead{szhao4@lenovo.com}
\author[inst1]{Weixiao Wang}
\ead{wwang29@lenovo.com}
\author[inst1,inst3]{Yaoman Li\corref{mycorrespondingauthor}}
\ead{ymli@cse.cuhk.edu.hk}
\author[inst3]{Irwin King}
\ead{king@cse.cuhk.edu.hk}

\affiliation[inst1]{organization={Lenovo Machine Intelligence Center},
            addressline={Hong Kong Science Park}, 
            country={Hong Kong}}
\affiliation[inst2]{organization={Delft University of Technology},
            city={Delft},
            country={The Netherlands}}
            
\affiliation[inst3]{organization={The Chinese University of Hong Kong},
            addressline={Shatin, NT}, 
            country={Hong Kong}}

\begin{abstract}
Meta-learning has gained wide popularity as a training framework that is more data-efficient than traditional machine learning methods. However, its generalization ability in complex task distributions, such as multimodal tasks, has not been thoroughly studied. Recently, some studies on multimodality-based meta-learning have emerged. This survey provides a comprehensive overview of the multimodality-based meta-learning landscape in terms of the methodologies and applications. We first formalize the definition of meta-learning in multimodality, along with the research challenges in this growing field, such as how to enrich the input in few-shot learning (FSL) or zero-shot learning (ZSL) in multimodal scenarios and how to generalize the models to new tasks. We then propose a new taxonomy to discuss typical meta-learning algorithms in multimodal tasks systematically. We investigate the contributions of related papers and summarize them by our taxonomy. Finally, we propose potential research directions for this promising field.
\end{abstract}


\begin{keyword}
Meta-Learning \sep Multimodal \sep Deep Learning \sep Few-shot Learning \sep Zero-shot Learning 
\end{keyword}

\end{frontmatter}

\section{Introduction}\label{intro}
Deep learning methods have made significant progress in the fields of speech, language and vision~\cite{DBLP:journals/ijon/GuoLOLWL16,DBLP:journals/access/ShresthaM19,DBLP:journals/cim/YoungHPC18,9737635}. However, the performance of these methods heavily relies on the availability of a large amount of labeled data which may be impractical or costly to acquire in most applications. To solve this problem, many researchers have actively explored two promising directions. One is applying the ``learning to learn" mechanism to gain or transfer knowledge from prior tasks to improve the learning efficiency in the new task. The other is obtaining heterogeneous modalities to enrich the model's input, e.g., instead of only looking at the image, feed related text description and the image itself to the model simultaneously. Recently, new state-of-the-art deep learning models have been proposed by modifying the meta-learning algorithm in multimodal scenarios. \par 
The ``learning to learn" mechanism~\cite{DBLP:books/sp/98/ThrunP98} used in the human learning process enables us to quickly learn new concepts from very few samples~\cite{DBLP:conf/nips/XingROP19}. Existing evidence has shown that humans can gain experience on multiple prior tasks over bounded episodes by combining prior knowledge and context. The learned abstract experience is generalized to improve future learning performance on new concepts. Inspired by this, a computational paradigm called meta-learning~\cite{DBLP:journals/corr/abs-2004-05439,DBLP:journals/access/KhanZRA20} is proposed to simulate the ability of humans to learn generalized task experience. Meta-learning allows machines to acquire prior knowledge from similar tasks and quickly adapt to new tasks. In addition, the process of extracting cross-domain task goals in a dynamic selection~\cite{DBLP:journals/air/VilaltaD02,DBLP:conf/iclr/RusuRSVPOH19} makes the meta-learning process more data-efficient than traditional machine learning (ML) models.\par
Due to the ability of meta-learning to generalize to new tasks, we aim to understand how meta-learning plays a role when the tasks are more complicated, for example, the data source of the task is no longer unimodal. Recent studies have focused on applying the meta-learning framework to the distribution of complex tasks~\cite{sikka2020multimodal,DBLP:conf/nips/VuorioSHL19,DBLP:conf/iconip/LiK20}, but are limited to a single modality. In particular, meta-learning has proven to be successful in both multi-task and single-task scenarios spanning various applications~\cite{DBLP:journals/corr/abs-2004-05439}, learning prior knowledge on the optimization steps~\cite{DBLP:conf/icml/FinnAL17}, data embedding~\cite{DBLP:conf/nips/SnellSZ17,DBLP:conf/nips/VinyalsBLKW16} or the model structure~\cite{DBLP:conf/iclr/RaviL17}. However, given the heterogeneous task modalities, the smart usage of meta-learning presents researchers with unique challenges. Each instance in the multimodal scenarios consists of two or multiple items of the same concept but different modalities. \par 
Related research on meta-learning has explored limited approaches to manipulate multiple modalities. The heterogeneous features of different modalities are first explored in ZSL/generalized zero-shot learning (GSZL) for image classification. Semantic modalities provide prior solid knowledge and assist visual modalities in model training, while meta-based algorithms are widely introduced to transfer knowledge more effectively from seen classes to unseen classes to capture the attribute relationships between paired modalities. However, the training process mainly regards one modality as the primary modality and exploits additional information by adding another modality. It does not involve the analysis of multiple modalities in real complex scenarios such as unpaired modalities, missing modalities, and associations between modalities. \par
Therefore, some studies have gone further to apply the meta-learning methods to tasks composed of other modality patterns. The data in some tasks might be missed or unbalanced. Since the modalities of different tasks usually come from different data distributions, the strengths of different modalities could be integrated into the problem to boost performance by taking full advantage of the multimodal data in the context of meta-learning. In the meantime, the training framework of meta-learning can also help improve the generalization ability of the original multimodality-based learner in new tasks. \par
Existing surveys only discuss meta-learning or multimodal learning separately, resulting in a research gap of the combination of multimodality with meta-learning. This paper provides the first systematic and comprehensive investigation of multimodality-based meta-learning algorithms. Our contributions can be summarized as:
\begin{itemize}
\itemsep0em
    \item proposing a new taxonomy for multimodality-based meta-learning algorithms and providing insightful analysis for each category;
    \item identifying the challenges in the new promising multimodality-based meta-learning area;
    \item summarizing specific works to solve different challenges, including their methodologies and differences;
    \item highlighting current research trends and possible future directions.
\end{itemize}
\par
The rest of the survey is organized as follows. We first briefly introduce the historical development of meta-learning and multimodality, formalizing their definitions. Then, we give the overall paradigm of challenges of multimodality-based meta-learning in Section \ref{section-foundations}. We propose a new taxonomy based on prior knowledge that meta-learning algorithms can learn in Section \ref{section-tax}. We respectively investigate the relevant researches of how to adapt the original meta-learning methods to multimodal data in Section \ref{section-opt}, Section \ref{section-emb} and Section \ref{section-data}, followed by a summary of these works in Section \ref{section-overview}. Finally, we conclude the paper with current research trends in Section \ref{section-dis} and possible directions for future work in Section \ref{section-fut}.

\section{Foundations}\label{section-foundations}
\subsection{A Historical Perspective}
\begin{figure}
    \centering
    \includegraphics[width=\columnwidth]{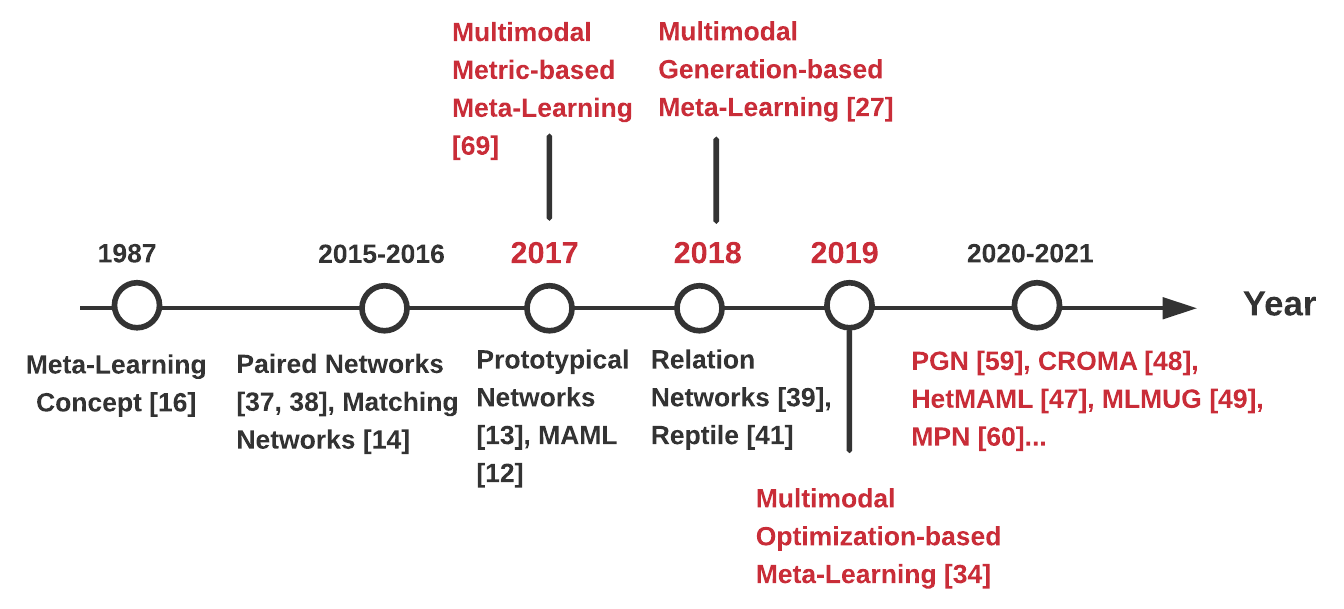}
    \caption{Chronological view of partial multimodality-based meta-learning methods. The year with \textbf{\textcolor[RGB]{201,45,57}{red}} indicates the first introduction of multimodality for the corresponding model. Methods with \textbf{\textcolor[RGB]{51,51,51}{black}} indicate traditional meta-learning models. Methods with \textbf{\textcolor[RGB]{201,45,57}{red}} all belong to the category of multimodality-based meta-learning.}
    \label{timeline}
\end{figure}

The concept of meta-learning first appeared in an article describing how to ``learn to learn''~\cite{schmidhuber:1987:srl}. It was further extended by the discovery of using evolutionary algorithms~\cite{bengio1990learning} to help learn the rules. More detailed methods only started to flourish after gradient descent and backpropagation were introduced to train meta-learning in 2001~\cite{DBLP:conf/icann/HochreiterYC01,younger2001meta}. Table \ref{ml-surveys} briefly shows the contributions of recent surveys to meta-learning. As an alternative paradigm to further promote the performance of ML with low-shot data, meta-learning has been shown to improve the generalization ability of computing models~\cite{DBLP:journals/corr/abs-1907-07287,DBLP:conf/icml/SantoroBBWL16}. Even though ZSL/GZSL algorithms for image classification have explored auxiliary modalities in the multimodal embedding space, the most common improvement is still applied on the unimodal task sets~\cite{DBLP:conf/icml/FinnAL17,DBLP:conf/nips/SnellSZ17,DBLP:conf/nips/KhodadadehBS19,DBLP:conf/isnn/LiK19}. Multimodal research first appeared in the exploration of audiovisual interaction in the process of speech perception~\cite{DBLP:journals/cm/YuhasGS89}. Later, with the massive increase of digital multimedia content, content analysis for multimodal sources became increasingly popular, along with combining deep neural networks~\cite{DBLP:conf/icml/NgiamKKNLN11} in applications such as visual question-answering~\cite{DBLP:conf/emnlp/FukuiPYRDR16} and image captioning~\cite{DBLP:journals/tcsv/YuLYH20}. The promising performance of multimodal learning models for visual semantic interaction is largely reflected in the ability to infer the potential alignment of images and semantics between modalities. However, this methodology cannot be directly applied to the case of FSL/ZSL. The pursuit of multimodality shows that training data with one modality can benefit from the alignment of different modalities~\cite{DBLP:journals/corr/abs-1806-05147}, while maintaining robust knowledge transfer between unseen and seen classes. Figure \ref{timeline} displays the historical development of meta-learning methods and highlights several time frames for the first introduction of multimodality.

\begin{table}[h]
    \scriptsize
    \centering
    \caption{Recent meta-learning survey contributions between 2010 and the present.}
    \begin{tabularx}{\columnwidth}{cccccc}
    \hline
    Year & Reference & Taxonomy & Architecture & Trends & Applications \\
    \hline
    2012 & \cite{bhatt2012survey} & \ding{55} & \ding{55} & \ding{51} & \ding{55} \\
    2015 & \cite{DBLP:journals/air/LemkeBG15} & \ding{55} & \ding{51} & \ding{51} & \ding{55}\\
    2018 & \cite{DBLP:journals/corr/abs-1810-03548} & \ding{51} & \ding{51} & \ding{51} & \ding{55}\\
    \multirow{5}{*}{2020} & \cite{DBLP:journals/corr/abs-2004-05439} & \ding{51} & \ding{55} & \ding{51} & \ding{51}\\
    & \cite{DBLP:journals/csur/WangYKN20} & \ding{51} & \ding{55} & \ding{55} & \ding{51} \\
    & \cite{DBLP:journals/access/KhanZRA20} & \ding{51} & \ding{51} & \ding{51} & \ding{51}\\
    & \cite{DBLP:journals/corr/abs-2004-11149} & \ding{51} & \ding{51} & \ding{51} & \ding{51} \\
    & \cite{DBLP:journals/corr/abs-2007-09604} & \ding{55} & \ding{55} & \ding{55} & \ding{51} \\
    \multirow{2}{*}{2021}& \cite{DBLP:journals/air/HuismanRP21} & \ding{55} & \ding{51} & \ding{51} & \ding{51} \\ 
    & \cite{DBLP:conf/icccnt/DokeG21} & \ding{55} & \ding{55} & \ding{55} & \ding{51}\\
    \hline
    
    \end{tabularx}
    \label{ml-surveys}
\end{table} 

\begin{figure*}[ht]
    \centering
    \includegraphics[width=0.9\textwidth]{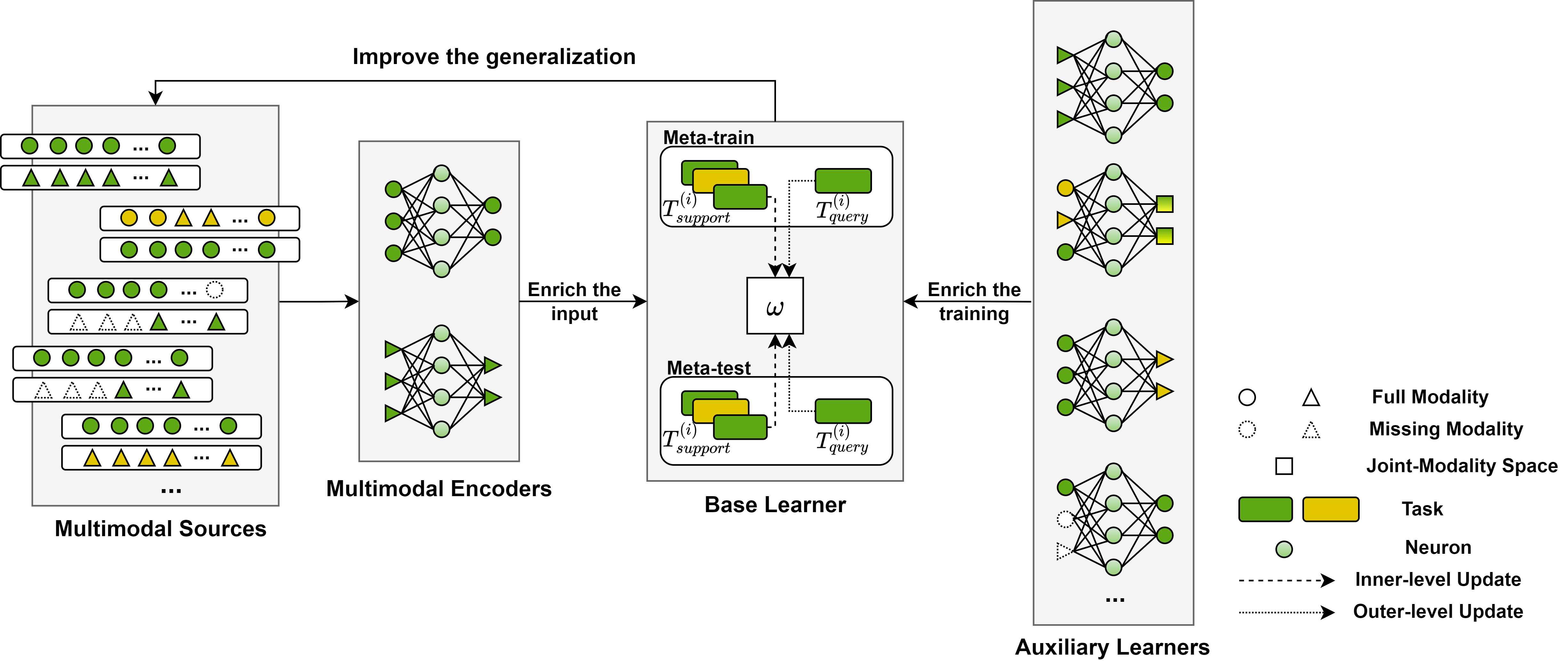}
    \caption{Paradigm of multimodality-based meta-learning. (1) Given the task segmentation of a multimodal dataset, we first list some possible task combinations with different modalities, including same tasks with same modalities, different tasks with mixed modalities, same tasks with missing modalities, different tasks with different modalities, etc. (2) Each modality needs to be encoded with the corresponding multimodal encoder as an enrichment to the input of subsequent models. (3) The base learner represents the main network trained by any meta-training algorithm that learns the meta-parameter $\omega^*$. The dataset is split for the meta-train and meta-test stage, in the form of support set $T^{(i)}_{support}$ and query set $T^{(i)}_{query}$. (4) Auxiliary learners are provided to tackle the multimodality-related problems, such as generating conditional modalities, mapping heterogeneous task distributions to the joint-modality space, aligning cross-modal tasks, reconstructing the missing modalities, etc. These learners are integrated with the base learner to enrich the original meta-training process. (5) As a result, the multimodality-based meta-learning paradigm finally improves the generalization across tasks from multimodal sources.}
    \label{paradigm}
\end{figure*}

\subsection{Formalizing Meta-Learning}
A common perspective of meta-learning is to learn algorithms that can be generalized across tasks. Following the definition in~\cite{DBLP:journals/corr/abs-2004-05439}, we denote the set of $M$ source tasks used in the meta-training stage as $D_{source}=\{(\mathcal{D}_{source}^{train},\mathcal{D}_{source}^{val})^{(i)}\}_{i=1}^{M}$ where each task is trained on $\mathcal{D}_{source}^{train}$ to minimize the internal loss function $\mathcal{L}^{task}$ in Eq.~(\ref{eq:lower}). Then the meta-knowledge $\omega^*$ is learned when evaluating on $\mathcal{D}_{source}^{val}$ over all the source tasks: 
\begin{align}
 \omega^*&=\mathop{\arg\min}_{\omega} \sum^{M}_{i=1}\mathcal{L}^{meta}(\theta^{*~(i)}(\omega), \omega, \mathcal{D}^{val~(i)}_{source})\label{eq:upper},\\
 \text{s.t. }\theta^{*(i)}(\omega) &= \mathop{\arg\min}_{{\theta}}\mathcal{L}^{task}(\theta, \omega, \mathcal{D}^{train~(i)}_{source})\label{eq:lower}. 
\end{align}
Usually, $\mathcal{D}_{source}^{train}$ and $\mathcal{D}_{source}^{val}$ are called support sets and query sets respectively. For the set of $Q$ target tasks in meta-test stage $D_{target}=\{(\mathcal{D}_{target}^{train}, \mathcal{D}_{target}^{test})^{(i)}\}_{i=1}^{Q}$, the learned prior knowledge $\omega^*$ in the meta-training stage will be used to train the model on unseen tasks. 
\par
The meta-learning paradigm discussed in this survey has two key properties: 
\begin{itemize}
    \item The existence of a task segmentation mechanism. The meta-training and meta-test sets~\cite{DBLP:journals/corr/abs-2004-05439} contain task units, where each task is divided into the support set and query set. Prior knowledge accumulated in the meta-training stage is used in the meta-test to evaluate the accuracy of the meta-model.
    \item The process of constructing an adaptive learner. This refers to the accumulation of meta-knowledge experience by dynamically improving the deviation between the inner base learner and the outer generalized learner. The type of meta-knowledge could be the estimation of initial parameters~\cite{DBLP:conf/icml/FinnAL17}, or an embedding strategy~\cite{DBLP:conf/nips/SnellSZ17}. Ultimately, it guarantees that the adapted knowledge can directly generalize to predicting new tasks instead of adding extra data. 
\end{itemize}

\subsection{Formalizing Multimodality}
We consider data modalities to be multiple when data is collected through multiple sensors, measurement equipment or acquisition technologies~\cite{ramachandram2017deep}. The output of each sensory channel is represented as a modality, which is usually a single form of dataset associated with a medium of expression, such as vision data from seeing objects and audio data from hearing sounds. The key characteristics constituting multimodality that cover data coming from multiple modalities can be summarized as follows:
\begin{itemize}
\itemsep0em
    \item Complementarity. Each modality brings a certain type of added value to the whole, and these added values cannot be inferred or obtained from any other modality in the setting~\cite{lahat2015multimodal}. 
    \item Diversity. The meaning is composed of different modalities from different semiotic resources.
    \item Integrity. The making of meaning involves the overall attention to the potential and limitations of each modality~\cite{jewitt2016introducing}. 
\end{itemize}

\par
Our key assumption is to combine multimodal features, such as images, fine-grained descriptions, auxiliary audio and video information to force the model to recognize cross-modal discriminative features to promote usage in meta-learning applications. Under this assumption, we limit the scope of our research to models that use multiple sensors as inputs, thereby excluding studies that build their models on multimodal task distributions with disjoint modes~\cite{sikka2020multimodal,DBLP:conf/nips/VuorioSHL19} or shifted domains~\cite{DBLP:conf/icml/KohSMXZBHYPGLDS21}.  \par
On the other hand, there are various combinations of multimodal sources in different tasks (See Figure \ref{paradigm}). Modalities could be provided in pairs or across tasks. The diversity and integrity of modalities are therefore reflected in the dependence on the semantics of different modalities. New classes with insufficient training data in unimodality can benefit from previously learned features. When one of the modalities is absent, the complementarity between modalities provides a way to conditionally consider auxiliary semantics and support the global meaning. 
\subsection{Challenges of Multimodality-Based Meta-Learning}
As a computing paradigm, meta-learning mainly learns shared meta-knowledge among different tasks and expresses it by transferring knowledge from seen classes to unseen classes. Integrating this framework with multimodality involves (i) using multimodal properties to enrich the task input of the meta-based model, and (ii) using auxiliary multimodality-based learners to enrich the task training of the meta-based model. Figure \ref{paradigm} illustrates the overall paradigm. \par

For problem (i), the research challenge is to make the learned meta-knowledge incorporate values from multimodal sources rather than a single modality. At the same time, the generalization ability of meta-knowledge in new tasks should also exceed the performance in unimodal situations. Different modalities have been encoded into relevant embeddings before being fed into the meta-learning network. Therefore, the base learner does not need to care about the specific representation of the modalities but can focus on how to use existing algorithms to adapt to multimodal inputs, including whether to use multiple modalities separately or to use fused modalities, how to combine modalities wisely and so on. In addition, there are scenarios where the meta-learning framework is only employed as an episode-based training method which could facilitate training efficiency. When multimodal inputs are added to such applications, the challenge becomes how to use the meta-learning framework to divide the multimodal task sets to achieve the ability of generalization. \par

For problem (ii), auxiliary tasks originating from multimodal problems have brought extra challenges to train the auxiliary learners and the base learner in a meta-learning manner. Common practices in multimodal ML~\cite{DBLP:journals/pami/BaltrusaitisAM19} address how to represent, summarize, map, align and fuse multimodal data in a way that exploits the characteristics of multiple modalities. When we integrate these methods with the base learner, auxiliary learners must participate in the co-training process to deal with the multimodal challenges. In other words, the essence of meta-learning then becomes how to provide a parameterized training framework, which means that the addition of these auxiliary networks needs to establish a relationship with the training parameters of the base network before they can be jointly trained. Hence, it is non-trivial to enrich the original meta-training process by adapting the inner-level update and the outer-level update to new networks that incorporate multimodal training, thus providing extra meta-knowledge about the prior modalities.\par
Furthermore, although we discuss the influence of modalities on the meta-learning framework from a multimodality-based perspective, the influence between these two concepts is mutual. Since the meta-learning training framework can help traditional multimodal learning to better generalize to new tasks, additional challenges may also include elements that affect the effectiveness of generalization, such as the definition of multimodal tasks, the choice of meta-learning algorithms, and the number of modalities. 

\begin{figure*}[ht]
    \centering
    \includegraphics[width=0.9\textwidth]{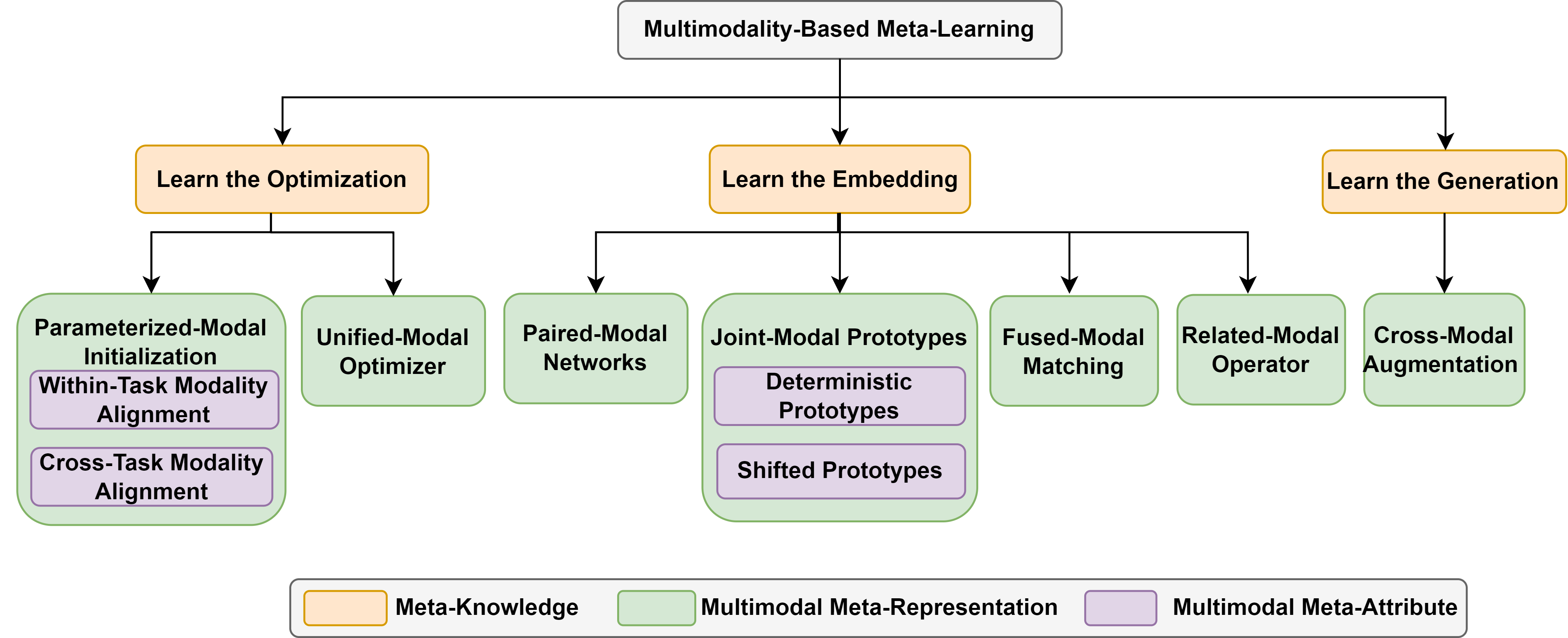}
    \caption{Taxonomy of multimodality-based meta-learning.}
    \label{taxonomy}
\end{figure*}
\section{Proposed Taxonomy}\label{section-tax}
Recently, many categorizations of meta-learning algorithms have been proposed~\cite{DBLP:journals/corr/abs-2004-05439,DBLP:journals/air/VilaltaD02,DBLP:journals/air/HuismanRP21} to explore what meta-knowledge can be learned. However, they mostly refer to application scenarios of the algorithms and ignore the influence of input data sources on the algorithms. To bridge the gap in the interaction between the input of multimodal sources and the form of algorithms in the current research, we are the first to introduce a new multimodality-based meta-learning taxonomy. We still establish the taxonomy on the type of meta-knowledge that can be learned, due to the distinct training methods employed by meta-learning algorithms. \par
In contrast to previously proposed categorizations, we discuss the algorithm details after combining with multimodal characteristics and the specific adjustments made for different multimodal alignments. Our taxonomy is based on three types of meta-knowledge: learn the optimization, learn the embedding, and learn the generation. For each meta-knowledge, we further provide a specific subcategory that reflects the influence of different multimodal meta-representations on the algorithms. When the relevant algorithms involve different variants operating on multimodal data, we will further discuss patterns presented in these variants as meta-attributes. Figure \ref{taxonomy} shows the taxonomy.\par
\textbf{Learn the Optimization} 
Optimization-based meta-learning methods focus on learning generalized outer-level parameters across tasks~\cite{DBLP:journals/corr/abs-2004-05439,DBLP:journals/air/VilaltaD02}. In a unimodal application scenario, the optimization usually only involves the main task such as the refinement of the base network structure~\cite{DBLP:conf/icml/FinnAL17}, without fine-tuning possible parameters in auxiliary tasks. When applying the learning paradigm on multimodal data, such multimodality needs to be parameterized before being trained along with the base network. Mathematically, the inner-level optimization of Eq.~(2) can be modified as:
\begin{align}\label{optimization-math}
    \theta_{m}^{*(i)}(\omega) &= \mathop{\arg\min}_{{\theta_{m}}}\mathcal{L}^{task}(\theta_{m}, \omega, \mathcal{D}^{train~(i)}_{source}), 
\end{align}
where $\theta_{m}$ denotes the collection of multimodal parameters, which are supposed to be diverse, such as the relationship of different modalities, the joint distribution of multiple modalities, and the weights of missing modalities.
\par
As far as we know, current optimization-based methods are dedicated to exploring how to use the meta-learning framework to help modify the training paradigm of initial parameters that define the modalities~\cite{DBLP:conf/www/YaoLWTL19} or relevant auxiliary networks~\cite{DBLP:conf/aaai/MaRZTWP21,DBLP:journals/corr/abs-2105-07889}. We, therefore, summarize the learnable meta-representations as parameterized-modal initialization and unified-modal optimizer. To capture how modality parameters are optimized in different task distributions, within-task modality alignment and cross-task modality alignment will be discussed in Section \ref{section-para}.
\par
\textbf{Learn the Embedding}
Metric-based meta-learning methods are widely used as non-parametric algorithms for few-shot problems. The idea is to learn an embedding network that helps project the training and testing points onto the same space to implement similarity comparison. The embedding network is trained on unimodal support sets where the quality of feature embeddings is often restricted. The introduction of multimodality expands the number of original embedding spaces. We emphasize that the embedding network aligning multiple spaces or adapting to the selection of multiple spaces can be applied to match any source of multimodal information, with the ability to capture the semantic relationship between input pairs more effectively. 
\par
Here, we summarize four widely used approaches originated from Pairwise Networks~\cite{koch2015siamese,DBLP:journals/corr/HofferA14}, Prototypical Networks~\cite{DBLP:conf/nips/SnellSZ17}, Matching Networks~\cite{DBLP:conf/nips/VinyalsBLKW16} and Relation Networks~\cite{DBLP:conf/cvpr/SungYZXTH18}. For each one in the multimodality-based scenario, modalities are respectively provided with pairs for the pairwise networks, combined to train a new joint prototype, fused to formulate the attention kernel, and related together to form the concatenation for the non-linear operators. Especially for variants of Prototypical Networks, prototypes play different roles in the model due to the different stages of multimodal fusion. Details about the difference between deterministic prototypes and shifted prototypes will be discussed in Section \ref{section-prototype}. \par
\textbf{Learn the Generation}
Another meta-knowledge is data generation, which is widely employed for most unimodal few-shot and zero-shot applications. Since the source of unimodal data is relatively simple and scarce, we exploit prior knowledge of the data to modify the model or execute data augmentation. There are often implicit relationships between multiple modalities, which can help better describe the scarce label characteristics across tasks. Although not many, we notice that some researchers have worked on the knowledge of conditional probability to simulate distribution of the primary modality conditioned on the auxiliary modality~\cite{DBLP:journals/corr/abs-1806-05147,DBLP:conf/aaai/VermaBR20}. We summarize the related methods as cross-modal augmentation with generating data across modalities.
\section{Learn the Optimization}\label{section-opt}
Optimization-based meta-learning methods aim to solve the inner-level task as an optimization problem~\cite{DBLP:journals/corr/abs-2004-05439} and extract the meta-parameter $\omega^*$ through Eq.~(\ref{eq:upper}) to obtain the best performance for generalizing to new tasks within only a few gradient-descent updates. Most of these methods are built on top of the Model-Agnostic Meta-Learning (MAML)~\cite{DBLP:conf/icml/FinnAL17} framework, which is a popular approach to meta-learn an initial set of neural network weights adapted for fine-tuning on few-shot problems. Reptile~\cite{DBLP:journals/corr/abs-1803-02999} generalizes the first-order MAML by repeatedly sampling a task and training to minimize the loss on the expectation over tasks. More alternatives that fit the meta-optimization process illustrated by Eq.~(\ref{eq:upper}) and Eq.~(\ref{eq:lower}) also learn specific inner optimizer by producing the trainable $\omega^*$ as its own hyperparameters~\cite{grefenstette2019generalized}. The exact optimization algorithm~\cite{DBLP:conf/iclr/RaviL17} will then generalize to unseen tasks to optimize the inner learner directly.\par
Multimodal information can usually be parameterized and wrapped into the inner-level optimization task, and then trained iteratively with the meta-parameter, either for the initialization or the optimization algorithm. We summarize the relevant studies in terms of where to use the meta-parameter and how it is co-trained with multimodal parameters. 
\subsection{Parameterized-Modal Initialization}\label{section-para}
\subsubsection{Overview}
Meta-learning algorithms such as MAML and Reptile are known to fine-tune parameters of unseen tasks quickly by using gradient descent methods. Since seen and unseen tasks are involved in the learning process across multiple domains, the diversity of the task space distributions puts forward higher requirements for applying the initialized meta-parameter. One common task distribution comprises tasks with the task instances of the same manifold from the same domain, where all tasks are either unimodal or multimodal. In addition, tasks may come from the same domain, but the task instances have different structures or subspaces~\cite{DBLP:journals/corr/abs-2105-07889}, such as using different modality features to describe the same semantic concept. We have reasons to believe that although the outer learner is still learning initial values that can be shared across different tasks, the generalization ability of these initialized parameters can be applied to more complex and multi-space task distributions. On the other hand, the heterogeneity gap~\cite{DBLP:conf/iccvw/SongT19} makes it more challenging to align multiple modalities into the same feature space across tasks. Therefore, according to the distribution pattern of parameterized multimodal information in the task space and how they are optimized to be aligned in the inner learner, we divide the current research into two branches, within-task modality alignment and cross-task modality alignment. \par
For the within-task modality alignment, the model's input comes from the homogeneous task space with different modality characteristics. Each task adopts the same type and number of modality feature spaces, such as paired modalities. The inner learner does not need to pay attention to the domain adaptation problems, but only needs to process the alignment of different modality subspaces~\cite{DBLP:journals/corr/abs-2105-07889}, such as reducing the dimensionality of some modalities, training a hybrid neural network for the mixed modality spaces, and constructing missing modality subspaces aligned with the known modalities~\cite{DBLP:conf/aaai/MaRZTWP21}. Finally, the meta-learning framework provides an iterative method of optimizing multiple parameters from multiple networks (i.e., inner network, outer network) together. Such a training framework still keeps the original structure of MAML or Reptile, in which the inner optimization is enriched by multiple modalities or other auxiliary training networks related to multimodal scenarios. \par
Multimodal tasks are usually sampled from a heterogeneous and more complex task distribution for the cross-task modality alignment. Each training task may contain a specific combination of modality subspaces. Most existing model-agnostic meta-learners assume that the tasks are evenly distributed, in which all tasks belong to the same domain and have the same manifold of task instances. However, such heterogeneous task distributions are often more challenging because instances with different task subspaces cannot share the same model structure completely. Therefore, related research examines the common and unique meta-parameters of different task spaces, exploring how to meta-align the knowledge of different subspaces to generalize to new tasks. The learned meta-knowledge $\omega^*$ is supposed to incorporate task-aware parameters as part of the initialization, where the task-specific information for each type of modality subspace may share the same manifold to adapt to the model-agnostic meta-learners.
 
\subsubsection{Methods}
\textbf{Within-Task Modality Alignment}
The parametric nature of the MAML framework enables multimodal problems to be characterized as the internal optimization involved in meta-training, where the inner network can be extended to multiple. Ma et al.~\cite{DBLP:conf/aaai/MaRZTWP21} focus on the implementation details by extending MAML to learn three networks in an integrated way for the main training process, missing modality reconstruction network and feature regularization by a Bayesian meta-learning based model, SMIL. There is no modality across different tasks involved obviously in training, but the use of a feature reconstruction network has leveraged the auxiliary modality to generate an approximation of the missing-modality feature efficiently. It has been highlighted in the paper that such a method is more efficient than the traditional generative methods (e.g., GAN, AE, or VAE) for not requiring full-modality data. The usage of single modality embeddings also guarantees that the approximation of full-modality is more flexible than the typical methods where feature spaces are regularized by the perturbation. Despite the successful collaboration training between modalities, an open challenge for deploying such Bayesian training framework in the real-world scenario is how to learn a set of proper modality priors for the unknown modalities.\par
\begin{figure*}[ht]
     \centering
     \begin{subfigure}[b]{0.44\textwidth}
         \centering
         \includegraphics[width=\textwidth]{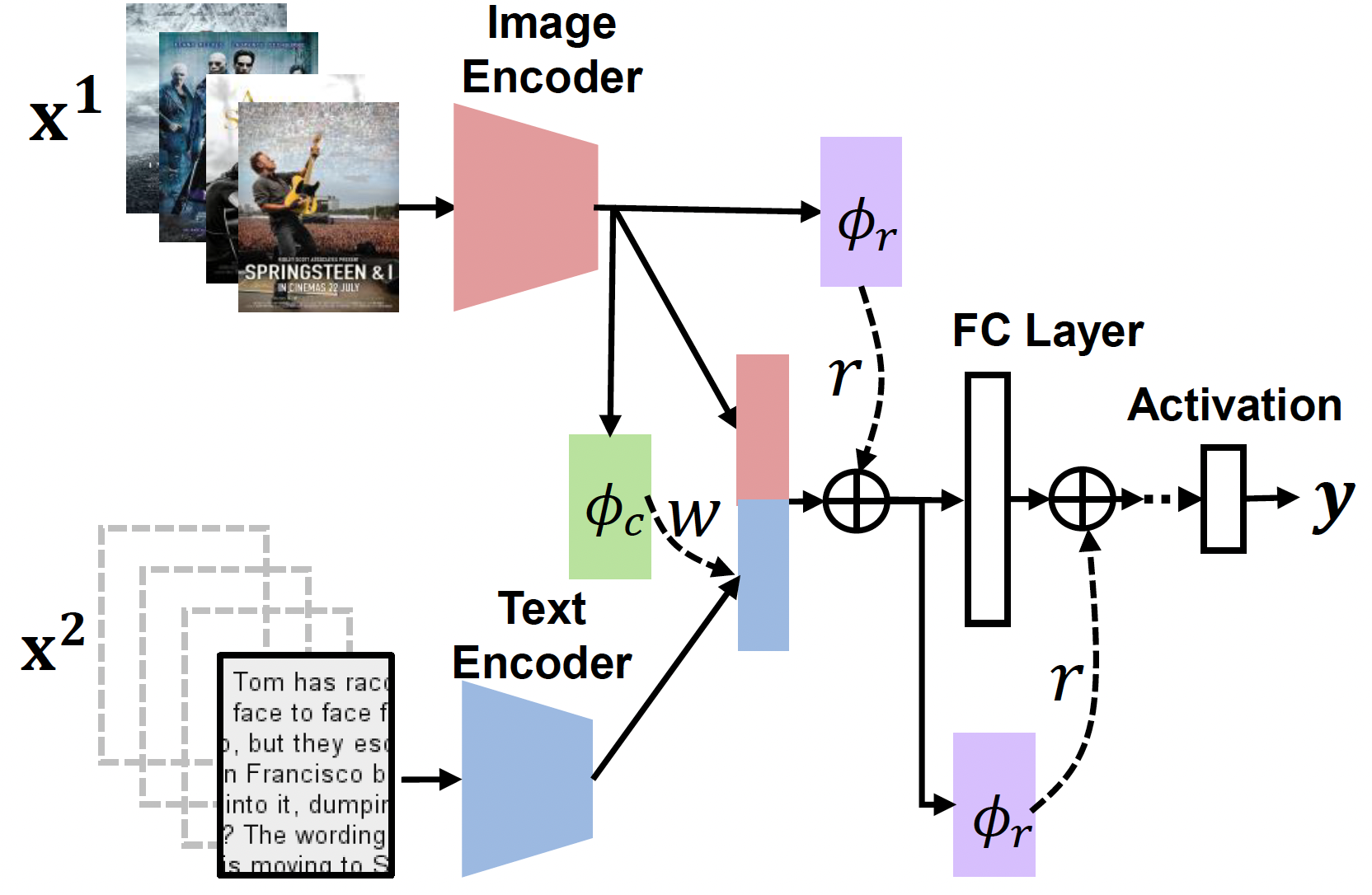}     
     \end{subfigure}
     \hspace{20pt}
     \begin{subfigure}[b]{0.44\textwidth}
         \includegraphics[width=\textwidth]{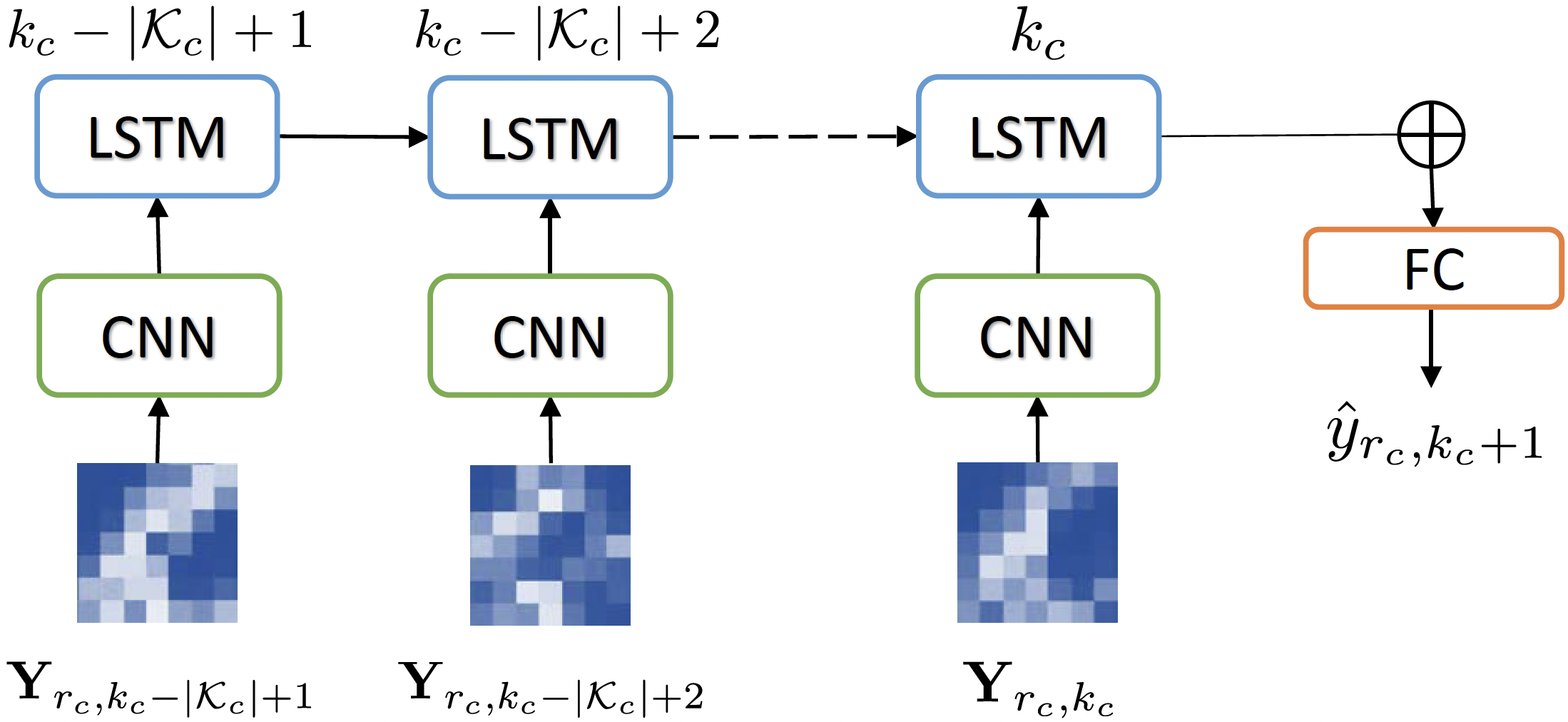}\vspace{4ex}
     \end{subfigure}
        \caption{Network structures for parameterized inner learners. \textbf{Left~\cite{DBLP:conf/aaai/MaRZTWP21}:} SMIL extends MAML by training the main network, the reconstruction network $\phi_c$, and the regularization network $\phi_r$ together. $x^i$ denotes different modalities. $\phi_c$ and $\phi_r$ together denote the inner optimization parameters while the main network is responsible for learning the meta-parameter $\omega^*$. \textbf{Right~\cite{DBLP:conf/www/YaoLWTL19}:} ST-net is parameterized by $\theta$ which encrypts the spatio-temporal correlations. $\theta$ denotes the overall parameters from CNN and LSTM where multimodal information is encoded. The initialization of $\omega$ from multiple source cities should be meta-learned.}
        \label{cotrain}
\end{figure*}

Moreover, the inner networks can be multimodal encoders for different features. Yao et al.~\cite{DBLP:conf/www/YaoLWTL19} try to reduce the risk of unstable results and knowledge transferred from a single source city. They propose a MetaST network in a meta-learning manner to learn to represent the initialization of spatio-temporal correlation based on tasks from multiple source cities. The estimated spatio-temporal values are computed by the hybrid model ST-net that combines a CNN and an LSTM to model each region's spatial dependency and temporal evolution. The parameters of the ST-net are updated in the inner-optimization to give representations of different modalities. Meta-parameters such as time dependence, spatial proximity, and regional function will be easily adapted to the target city by fine-tuning, and finally, serve for the relevant task predictions. MetaST has overcome the risk of transferring the knowledge from a single source and adapted the spatio-temporal sequences in various scenarios rather than only discrete features. However, the type of shared knowledge is still in the state of a black box, which poses a challenge to broader scalability. Figure \ref{cotrain} compares the above parameterized inner learners that could enrich the training of the meta-learning framework.
\par
In addition to the application of conventional multimodal task sets, the meta-learning framework can also help improve the training performance of neural network models in other real-world settings, such as helping agents in reinforcement learning (RL) to better adapt to different scenarios and helping solve the catastrophic forgetting of previous tasks in continual learning. \par
Specifically, Yan et al.~\cite{DBLP:conf/iros/YanLSY20} adopt the MAML framework in the experiment of indoor navigation from the perspective of RL. Tasks are divided based on various room scenes to help train and test the model using the aggregated features of visual and audio modalities. Different features of visual observations are encoded to aggregate the word embedding from audio features and then inputted to the memory network to carry out a sequence of action policies. The good initialization used for navigating different scenes is trained along with the inner parameters that control the optimization of the agent's interaction loss. It is evidenced that the meta-learning framework makes the model more robust to unseen scenes, but further validation on a real programmable robot is still missing. \par
Verma et al.~\cite{DBLP:journals/corr/abs-2102-11856} introduce a new meta-continual scenario to handle unseen classes that are collected in a sequential stream dynamically. While the main architecture is based on pairing self-gating of attributes and scaled class normalization, the need for balancing even generalization across all tasks during the reservoir sampling requires the model to be trained with Reptile in a meta-learning manner. Such an idea of avoiding expensive generative models has implications for applying multimodality-based meta-learning models in a real-world setting, where the one-time adaptation paradigm of the streaming data possibly fails. Similar work can be inspired to focus on catastrophic forgetting in the sequential data. \par
We can conclude that for the within-task modality alignment, the methods mainly focus on transforming modality issues into the design of auxiliary learners to enrich the training process. Flexible applications can be proposed as long as the involved multiple modalities can be parameterized and added to the gradient-descent updates.\par
\textbf{Cross-Task Modality Alignment}
The most common cross-task alignment is used to learn shared knowledge from different domain tasks. Modalities often appear in similar patterns in each task, but the semantic difference of different domains leads to the difficulty of transferring meta-knowledge across tasks. TGMZ~\cite{DBLP:conf/aaai/LiuLY0L21} addresses the limitations of meta-ZSL models that are mostly optimized on the same data distribution without explicitly alleviating representation bias and prediction mismatch posed by diverse task distributions. It uses an attribute-conditioned auto-encoder to align multiple task domains in a unified distribution with auxiliary textual modalities. Different from this study~\cite{DBLP:conf/aaai/VermaBR20} which searches optimal parameters for each task, the meta-training relies on the task-specific loss function to compute gradients on support and query sets for obtaining the overall optimal parameters of different modules, task encoder, task decoder, task discriminator, and task classifier. The performance of TGMZ over ZSL/GZSL has demonstrated the necessity of task alignment in alleviating discrimination in class distributions. This suggests a potentially robust way of training meta-ZSL models by speculating more about the disjoint task distribution. \par

\begin{figure*}[ht]
     \centering
     \begin{subfigure}[b]{0.44\textwidth}
         \centering
         \includegraphics[width=0.85\textwidth]{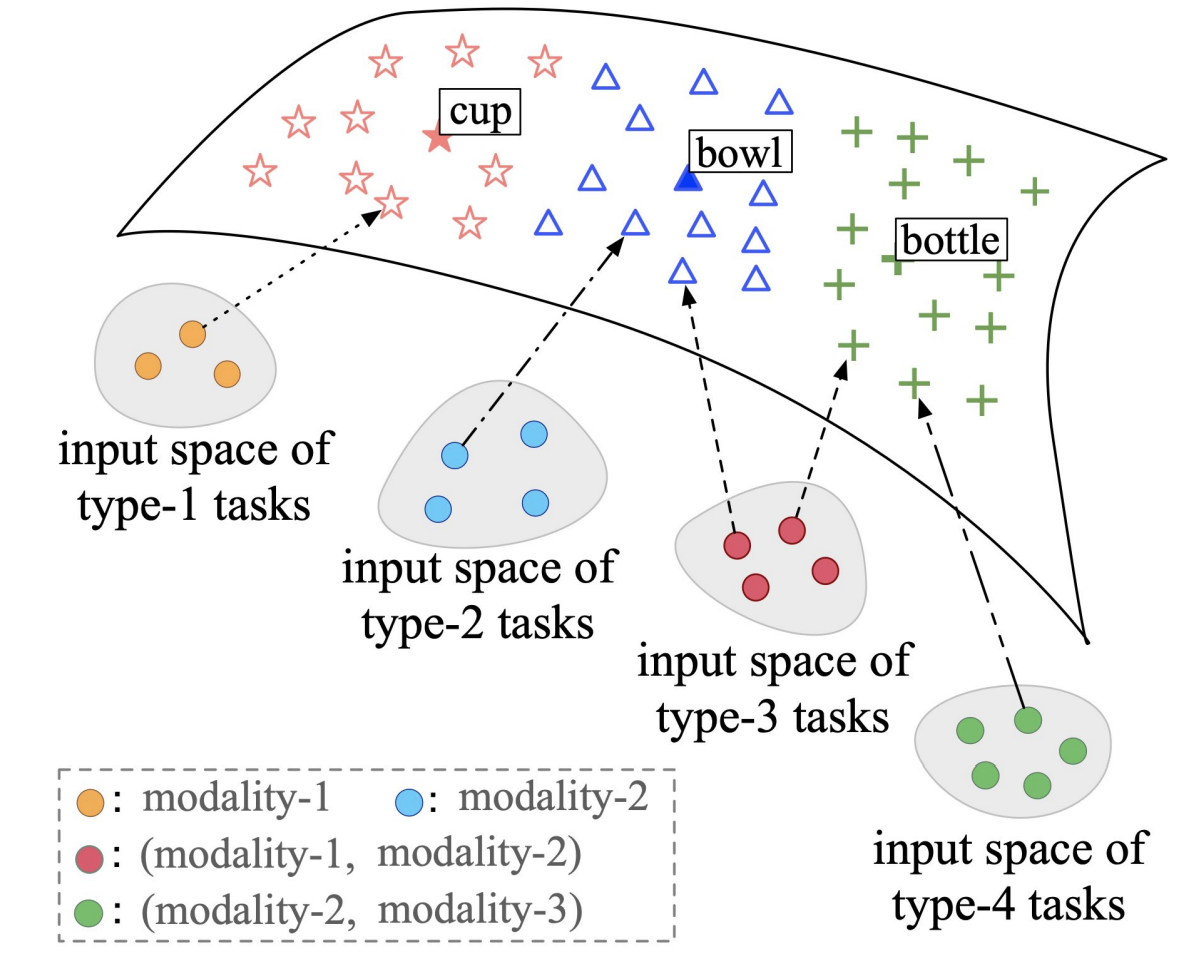}     
     \end{subfigure}
     \hspace{20pt}
     \begin{subfigure}[b]{0.44\textwidth}
         \includegraphics[width=0.85\textwidth]{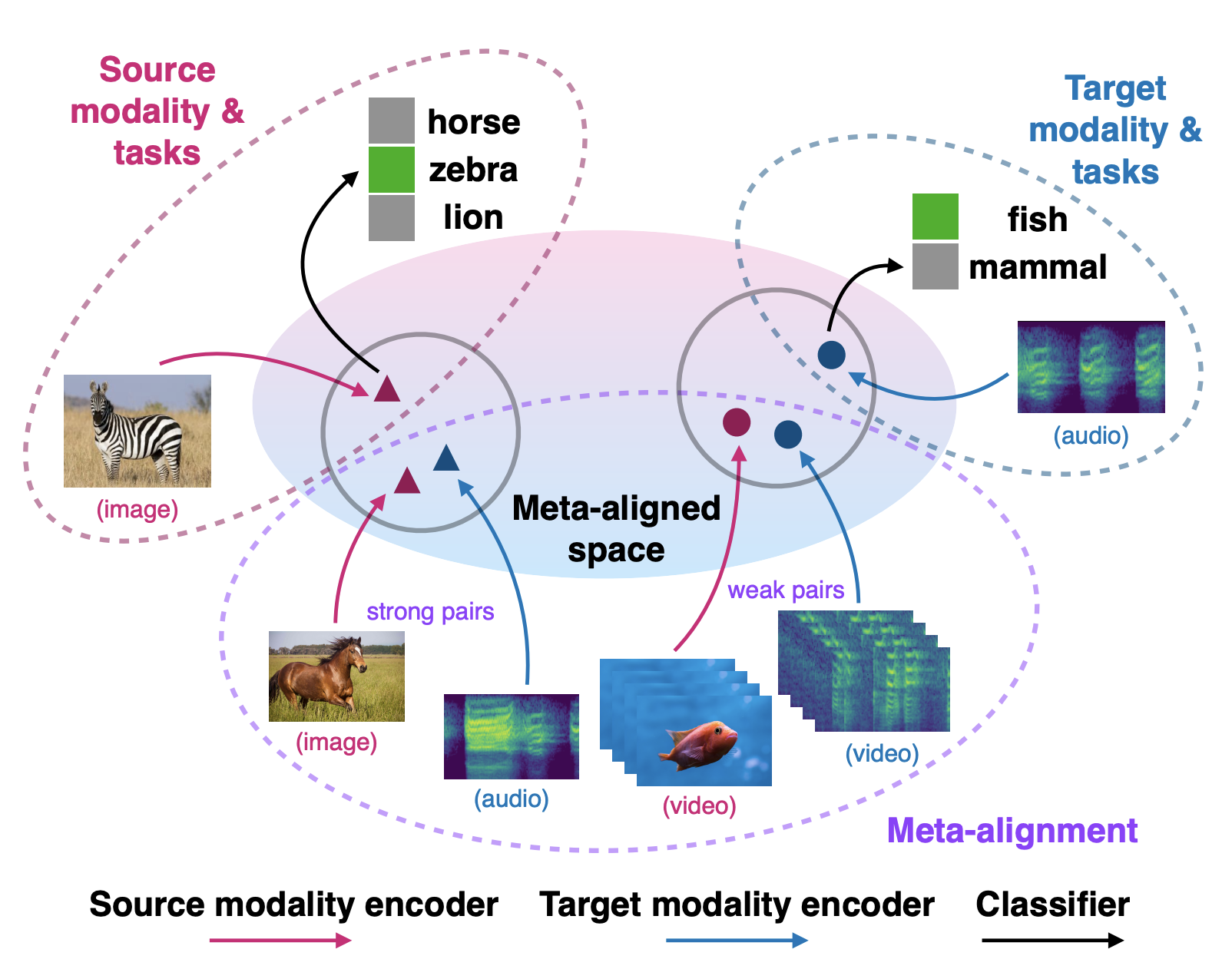}
     \end{subfigure}
        \caption{Meta-alignment representations of more generalized task distributions.  \ \textbf{Left~\cite{DBLP:conf/cikm/ChenZ21}: Heterogeneous task distributions.} Each task has its unique subspace describing the commonly shared concept domain. \textbf{Right~\cite{DBLP:journals/corr/abs-2012-02813}: Cross-modal task distributions.} Source tasks and target tasks have different modalities to be aligned.}
        \label{gener-tasks}
\end{figure*}
In addition to the existence of multi-domain labels in image classification applications, other multimodal scenes also include task sources from different domains. MLMUG~\cite{ma2021model} focuses on the user domain generalization issue that exists in cross-modal retrieval. To allow the model to generalize from source user domains to unknown user domains without any update or fine-tuning, the model first constructs a cross-modal embedding network to learn a shared modality feature space for cross-modal matching, which generates a shared mask used to encode transferable knowledge between different user domains. Then a meta-learning framework is implemented to learn the meta covariant attention module. Unlike the original MAML, the gradients from the meta-training set and meta-test set finally get weighted aggregation. MLMUG has provided a potential training paradigm for applications that cannot obtain unlabelled data instances from unknown fields in the real world, which possibly are more efficient than unsupervised domain adaptation. For applications with large domain spaces, this method gives access to generalized parameter learning instead of expensive joint training. \par
Despite the success of task-aware models in multiple task domains, different tasks are still limited to the same modality patterns. Another more generalized task distribution will be with different modality subspaces for different tasks. To align multiple subspaces, HetMAML~\cite{DBLP:conf/cikm/ChenZ21} tries to extend the application of gradient-based meta-models from homogeneous task distributions to heterogeneous tasks, where tasks in the same query set have different numbers of modalities or input structures. Compared with training different types of tasks separately, HetMAML aims to train a unified meta-learner that can simultaneously capture the global meta-parameters about the concept domain shared by all tasks, as well as the customized parameters that characterize each task. A three-module shared network architecture is proposed to achieve this goal, including multi-channel feature extractors, feature aggregations, and the head module for decision making. By adopting the BRNN structure, different modality-specific feature extractors are iteratively aggregated to enable all types of tasks to be mapped from heterogeneous feature space to a unified space. A better generalization of a few examples is not only reflected in the common meta-parameters shared between different types of tasks, but also in the type-specific meta-parameters for each task. HetMAML is a potentially powerful extension of the MAML framework, which not only retains the model-agnostic property of MAML, but embeds task-aware knowledge into the outer-level update.\par
Furthermore, CROMA~\cite{DBLP:journals/corr/abs-2012-02813} goes further to establish a more generalized meta-learning paradigm based on cross-modal alignment that is trained on different source modalities, and then quickly generalize to the target modality to perform new tasks. Specifically, they jointly train the cross-modal meta-alignment space and source modality classifier in the meta-training process to perform generalization on alignment and classification tasks by Reptile. The meta-parameters learned in the training process will be used as initial parameters for the meta-test process to classify tasks in the target modality. When the labeled data of the target modality is scarce, the joint space learned by the meta-alignment of the source modality and target modality will help transfer knowledge among tasks. Figure \ref{gener-tasks} compares the meta-alignment representations of heterogeneous task distributions and cross-modal task distributions. \par
When generalizing gradient-based meta-models to the cross-task modality alignment, the extent of what and how much knowledge across modalities should be shared has been heavily emphasized throughout the existing research. Both tasks from different domains or the same domain concept are parameterized and updated in the outer learning process to adapt to unknown tasks quickly. The major challenges of transferring knowledge under this framework remain open, such as implementing task-specific loss, the choice between modular and hierarchical network architectures, and the construction of the shared modality space. \par

\subsection{Unified-Modal Optimizer}
\subsubsection{Overview}
The meta-learning algorithm could learn the meta-optimizer by updating the inner learner at each iteration. The most typical one is the LSTM-based meta-learner~\cite{DBLP:conf/iclr/RaviL17}, aiming to use the sequential features of the LSTM model to simulate the cell states during the update process. The updating process approximates the gradient descent method, where the information stored in forget and input gates during training is improved. The parametrization of the optimizer allows new tasks to be learned from a series of previous tasks. When the multimodal training network is introduced, the meta-learner still keeps the idea of optimizing the optimizer to help update a better multimodal network, depending on the specific structure of the network. \par
When the initial value is meta-learned for multimodal tasks, the subspace of each modality needs to be learned together through an aligned common space. However, the modality input in the real world is often dynamically changing, and the adaptation of learning frameworks to different modalities under the sequential distribution is challenging. Recent work has employed the idea of human cognitive learning to address the adaptive training of multimodal models, where various modalities do not need to be aligned during training at a fine-grained level. Therefore, the LSTM-based meta-learner becomes an appropriate alternative to take advantage of the previous experience in the sequence to avoid catastrophic forgetting.
\subsubsection{Methods}
While most articles apply all the modalities to each task together, another way of thinking is to add the modalities sequentially, dynamically adjusting the performance of each task on the unified multimodal model. Ge and Xiaoyang~\cite{DBLP:conf/iccvw/SongT19} specifically propose the sequential cross-modal learning (SCML) from a novel perspective in terms of learning the optimization of the unified multimodal model. Each modality is learned sequentially by the unique feature extractor and then projected onto the same embedding space by the unified multimodal model. Then they extend the LSTM-based meta-learner~\cite{DBLP:conf/nips/AndrychowiczDCH16} to effectively optimize the new unified multimodal model based on the old experience well-trained on previous tasks. It is evidenced that due to the ability of the meta-learner, the updated unified model obtains a slight increase over the ordinary gradient descent method.\par
From the number of articles that use LSTM-based meta-learner, we can infer that the application of the LSTM model to deal with dynamic sequential multimodal data is still limited. While mapping multiple modalities to the same embedding space simultaneously has been adopted as the common practice, sequential learning of multiple modalities by the LSTM-based meta-learner is undoubtedly promising.   

\section{Learn the Embedding}\label{section-emb}
Metric learning approaches learn the non-linear embedding space, where intrinsic class memberships are decided by measuring the distances between points. The learned prior knowledge is often implemented as an embedding network or a projection function that transforms raw inputs into the representation suitable for similarity comparison~\cite{DBLP:journals/corr/abs-2004-05439} in a feed-forward manner~\cite{DBLP:conf/nips/SnellSZ17}. Popular embedding networks include Siamese Network~\cite{koch2015siamese}, Triplet Network~\cite{DBLP:journals/corr/HofferA14}, Matching Networks~\cite{DBLP:conf/nips/VinyalsBLKW16}, Prototypical Networks~\cite{DBLP:conf/nips/SnellSZ17} and Relation Networks~\cite{DBLP:conf/cvpr/SungYZXTH18}. Other advanced approaches such as graph-based networks~\cite{DBLP:conf/cvpr/KimKKY19,DBLP:conf/iclr/SatorrasE18} are also widely explored to model the relationship between samples. In addition, metrics such as Euclidean distance~\cite{DBLP:conf/wacv/LuoWLXP21}, cosine distance, contrastive loss, and triplet loss~\cite{DBLP:conf/iccv/ZhangZNXY19} are typically used to measure the similarities within pair or triplet samples. The flexible embedding network structures and metric computations provide convenience for introducing multimodality, which promotes the feature extraction and interpretation to be realized in various ways.
\subsection{Paired-Modal Networks}
\subsubsection{Overview}
Pairwise networks are introduced to take pair examples and learn their sharing feature space to discriminate between two classes. For example, Siamese Network~\cite{koch2015siamese} employs two identical neural networks to extract embeddings from a pair of samples and computes a weighted metric to determine the similarity. Triplet Network~\cite{DBLP:journals/corr/HofferA14} extends the networks to three with shared parameters to output the comparison probability. Although the pairwise comparators limit the possibility of training end-to-end networks directly for the few-shot problems, the learning framework is still valid to be applied to multimodal datasets. A common technique is to change the identical architectures that share parameters in the pairwise networks to different embedding networks for different modalities. New loss metrics are often proposed along with the networks to perform the matching task between inputs from different modalities. 
\subsubsection{Methods}
As the output of paired networks, a shared feature space allows the use of different modalities under the same distance metric. Liu and Zhang~\cite{DBLP:journals/corr/abs-2009-07879} propose STUM to take advantage of the siamese feature space formation process~\cite{DBLP:conf/cvpr/HadsellCL06} to learn a shared feature space by different networks for different modalities. Instead of using triplet loss during training, it also employs a simple summation of the contrastive loss function to handle the random number of streaming data representations coming from the time relation of inputs. Positive and negative samples group the inputs in one or more modalities that are adapted to non-identical networks for processing. STUM has effectively captured the formation process of the feature space within and across the modalities in time-cued data. The experiment on visual modalities has shown that the model can potentially be used for the feature organization of objects represented by multiple modalities. The high performance achieved without supervision also implies the potential extension to fast online learning.\par
\begin{figure*}[ht]
     \centering
     \begin{subfigure}[b]{0.44\textwidth}
         \centering
         \includegraphics[width=\textwidth]{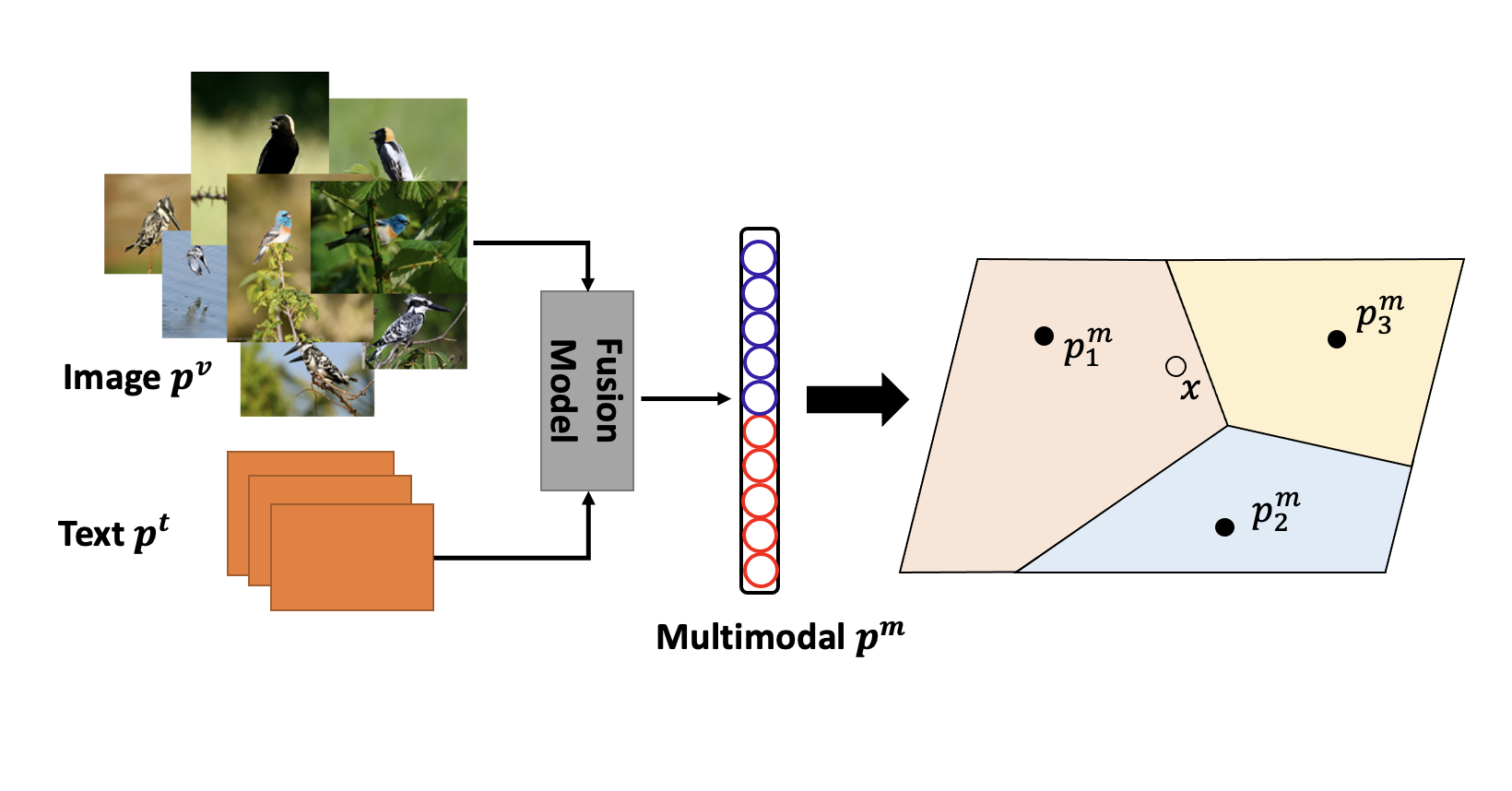}    
     \end{subfigure}
     \hspace{20pt}
     \begin{subfigure}[b]{0.44\textwidth}
         \includegraphics[width=\textwidth]{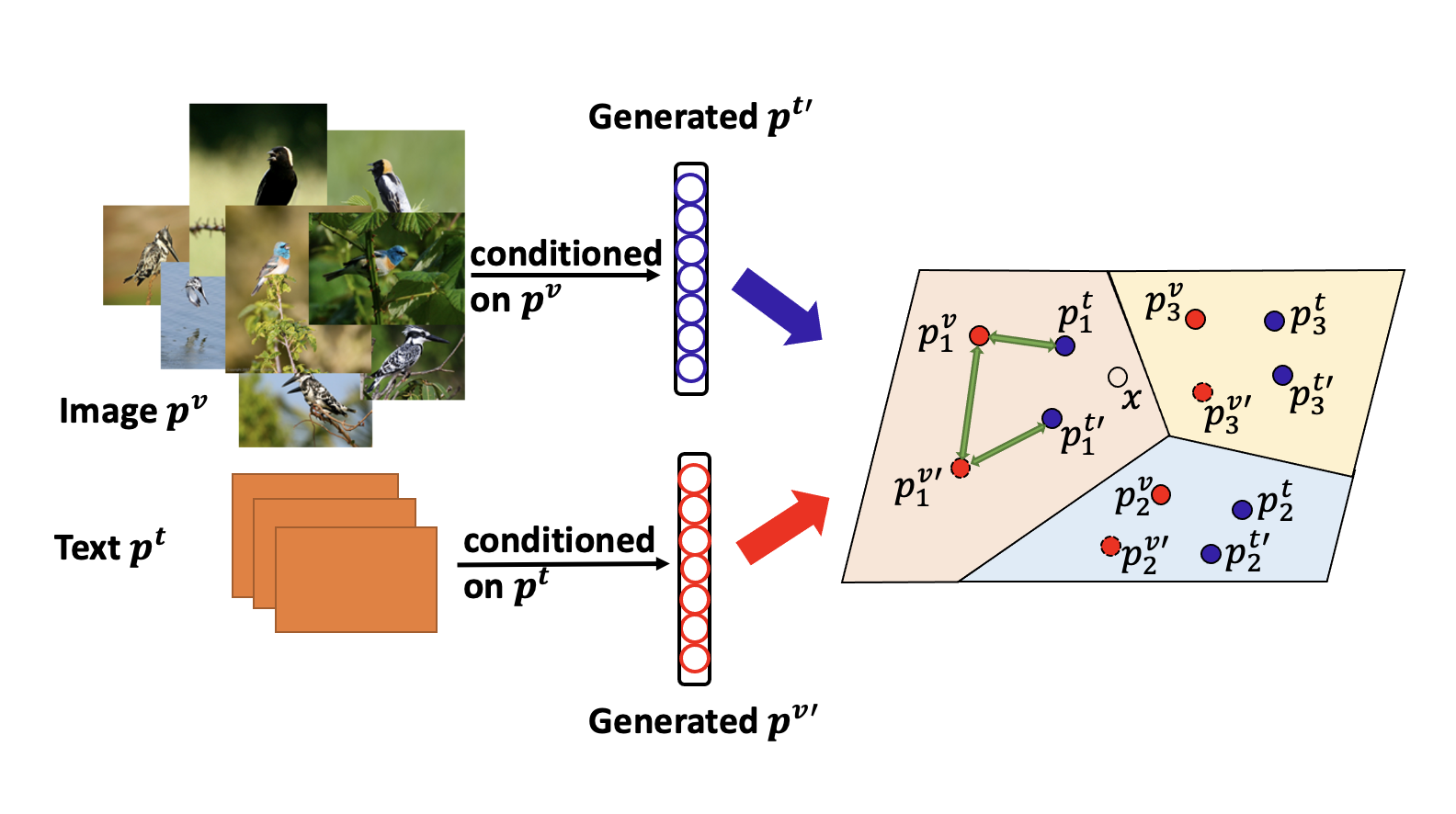}
     \end{subfigure}
        \caption{Representation of joint-modal prototypes. \textbf{Left}: \textbf{Deterministic prototypes}. Modalities are fused through the fusion model and transformed as the joint representation. Each multimodal prototype $p^m$ in each class is projected onto the multimodal embedding space. The query sample $x$ is then compared to different multimodal prototypes to obtain the prediction. \textbf{Right}: \textbf{Shifted prototypes}. New prototypes are generated by conditioning on the original prototypes in one modality. Flexible choices on original averaging prototypes and generated prototypes are introduced. Take one class $p_1$ as an example, AM3~\cite{DBLP:conf/nips/XingROP19} adaptively combines $p_1^v$ and $p_1^t$. Episode-based PGN~\cite{DBLP:conf/cvpr/YuJHZ20} modifies the objective function by integrating both of the generated prototypes $p_1^{v'}$ and $p_1^{t'}$. MPN~\cite{DBLP:conf/wacv/PahdePKN21} averages the conditioned $p_1^{v'}$ with the original $p_1^v$. The new prototype is supposed to be anywhere on the green line based on the value of the weighting factor.}
        \label{prototype}
\end{figure*}
Triplet loss is also often used as a loss variant in combination with paired networks to enhance the robustness of the model. Eloff et al.~\cite{DBLP:conf/icassp/EloffEK19} focus on learning a multimodal matching space from paired images and speech in the one-shot domain. They investigate a Siamese neural network built with the semi-hard triplet mining trick to alleviate the memory issue. Each spoken test query is matched to a visual set according to the learned embedding metric from training the support set. However, they do not approach the matching problem in a true multimodal space where the mapping happens mutually between two modalities, therefore transforming it into a two-step indirect comparison in the visual embedding space. \par
To learn a multimodal space in one step, Nortje and Kamper~\cite{DBLP:journals/corr/abs-2012-05680} improved upon the above idea to learn a shared embedding space of spoken words and images from only a few paired examples by optimizing the combination of two triplet losses. The model they came up with is MTriplet, which overcomes the weakness of learning representations that rely on unimodal comparisons despite having the support set of speech-image pairs. After preprocessing the two modalities by the corresponding autoencoders, a multimodal triplet network is learned to map inputs of the same class to similar representations measured by a direct cross-modal distance metric. \par
Even though paired-modal networks have demonstrated the potential for feature space formation, modality patterns extracted from real-world datasets remain an open challenge. When the number of modalities increases from two to multiple, it is unclear whether the network structure could achieve high performance under conditions such as missing modalities and scarce paired samples. \par
\subsection{Joint-Modal Prototypes}\label{section-prototype}
\subsubsection{Overview}
The term ``prototype'' refers to the centroid of each class within the dataset. In Prototypical Networks~\cite{DBLP:conf/nips/SnellSZ17}, a support set is used to calculate the prototype for each class, and the query samples are classified according to the distance from each prototype. The calculation of the prototype $p_c$ needs to rely on the averaged embeddings of all the support samples in class $c$ for each episode of training~\cite{DBLP:conf/nips/XingROP19}: 
\begin{equation}\label{prototype-repr}
   p_c = \frac{1}{|S_e^c|}\sum_{(s_i, y_i) \in S_e^c} f_{\theta}(s_i),   
\end{equation}
where $S_e^c \in S_e$ is the subset of support set belonging to class c, $S_e=\{(s_i, y_i)_{i=1}^{N\times K}\}$ and $f_{\theta}$ is the embedding network that needs to be learned. $S_e$ contains $K$ samples for each of the $N$ classes. In general, after acquiring the distances of the query embedding to the embedded prototype $p_c$ in each class, we can obtain the distribution of the query sample over all of the classes of the episode. The meta-objective of the model then becomes to minimize the expectation of the negative loglikelihood of the true class of each query sample~\cite{DBLP:conf/nips/XingROP19}:\par
\begin{equation}\label{loss}
L(\theta) = E_{(S_e, Q_e)} - \sum_{t=1}^{J}log p_{\theta}(y_t|q_t, S_e),
\end{equation}
where the query set $Q_e=\{(q_j, y_j)_{j=1}^{J}\}$ contains $J$ samples. $(y_t, q_t) \in Q_e$ and $S_e$ are the sampled query set and support set in each episode of training. \par
When the prototype embeddings are computed from more than one modality, we observe that the representation of prototypes could be modified according to the way of introducing different modalities. By exploiting whether modalities are fused before passing to the prototypical networks, we divide the representation of joint-modal prototypes into two categories, \textbf{deterministic prototypes} and \textbf{shifted prototypes}. \par
Deterministic prototypes use a unimodal vector extracted from the fusion of other models, which means usually different modalities are trained by different encoders, followed by the connection layer to concatenate the outputs after training. The prototypical networks are expected to use the concatenated unimodal vector to perform the following similarity calculation, so there is no obvious difference from the traditional prototypical networks. This method cares more about multimodal learning rather than the meta-learning framework, such as how to remove correlations between modalities and fuse them in a lower-dimensional common subspace. \par
By contrast, the early fusion of modalities may add noise to the embeddings since the information retained by different modalities is homogenized in this process. The idea of relying on learned embeddings and their distances to discriminate unseen classes in the prototypical networks has a similar problem structure to alignment, a key challenge in the field of multimodal learning. Alignment focuses on how to identify the direct relationship between elements in two or more different modalities~\cite{DBLP:journals/pami/BaltrusaitisAM19}. Existing research usually maps different modalities to the same semantic representation space, and then computes the similarity as a direct relationship measurement. The phenomenon of multimodal representations in the same semantic space allows extending the prototypical networks. \par
Relevant research often applies unimodal task training resulting in unimodal feature vectors of the prototypes. Accordingly, some studies have introduced multimodal feature vectors to shift the prototype representations. In that case, we define the method of shifted prototypes as the one that uses the various multimodal vectors directly without prior fusion and builds models upon the ensemble. Figure \ref{prototype} shows the two types of joint-modal prototypes with two commonly used modalities, image and text, as an illustration.\par

\subsubsection{Methods}
\textbf{Deterministic Prototypes.}
The computation of distances between deterministic prototypes in the joint space of different modality features follows the same approach as for traditional unimodal prototypes. The metric of the meta-learning algorithm does not require special modification. Wan et al.~\cite{DBLP:conf/aaai/WanZDHYP21} propose to learn the prototype for each class of the social relation extracting from the combination of facial image features and text-based features. The unbalanced distribution of social relations from real-life multimodal datasets motivates them to propose prototypical networks training on a random support set of tuples on different classes of social relations by using FSL techniques. A cross-modal encoder (Illustrated as the Fusion Model in Figure \ref{prototype} Left) then concatenates the normalized feature vectors of two modalities learned respectively from a pre-trained language model and FaceNet. Eventually, the prototypical networks are applied to predict the social relations in the query set by directly averaging concatenated cross-modal embeddings as the prototype for each class.\par
Deterministic prototypes are easy to obtain, which allows for accurately encoding different modalities before implementing the meta-learning framework. The early diffusion is also accessible to more modalities if the data pattern changes in the future. However, the effectiveness has not been justified theoretically, as most research does not favor deterministic prototypes but tends to use shifted prototypes.\par
\textbf{Shifted Prototypes.}
It is expected that leveraging auxiliary modality provides the means to inject diversity into the generated sample space for novel classes during training~\cite{DBLP:conf/cvpr/XianLSA18,DBLP:conf/cvpr/ZhuEL0E18,DBLP:conf/wacv/PahdeOJKN19}. Pahde et al.~\cite{DBLP:conf/wacv/PahdePKN21} aim to obtain more reliable prototypes and enrich the intrinsic feature sparsity in the few-shot training space. By training a generative model that maps text data into the pre-trained visual feature space, a new joint prototype $p_c^{'}$ for each novel class is redefined as the weighted average of both the original visual prototype $p_c$ and the prototype computed from the generated visual feature vectors $p_c^G$ conditioned on the text modality:
\begin{equation}\label{new-prototype}
    p_c^{'} = \frac{p_c+\lambda * p_c^G}{1+\lambda},
\end{equation}
\begin{equation}\label{condi-prototype}
    p_c^G = \frac{1}{|S_e^c|}\sum_{(s_i, y_i) \in S_e^c} G_t(\Phi_T(t_i)),
\end{equation}
where  $\lambda$ is a weighting factor, $G_t$ is a generator and $\Phi_T(t_i)$ is the textual representation. Compared to Eq.~(\ref{prototype-repr}), the embedding network $f_{\theta}$ is learned as the generator that generates new visual feature vectors. The method has verified that the combination of modalities allows the few-shot feature sparsity to be compensated and the class membership of novel samples to be better inferred.\par
Although the generative method has gained great success in the few-shot classification, it often assumes different modalities have the same information abundance and distinctiveness. In fact, for certain concepts, some modalities may be more discriminative than others due to the heterogeneous structure of their respective feature spaces. When the richness of different modalities does not match, the alignment is less useful. \par
Xing et al.~\cite{DBLP:conf/nips/XingROP19} propose an adaptive modality mixture mechanism (AM3) that can be applied to the context of image and text, retaining the independent structure of the two modalities. The model is built on top of the prototype network to augment metric-based FSL methods by incorporating the label embeddings of all categories in both the train set and test set. The two modalities can be adaptively used according to different scenarios based on a convex combination of the two modalities in the prototype representation of each category.\par 
Schwartz et al.~\cite{DBLP:journals/corr/abs-1906-01905} make a further improvement upon AM3 by adding multiple semantic explanations rather than just the category names to the convex combination of the visual and semantic prototypes. The introduction of multimodal prototypes and the pre-trained backbone models demonstrate the effectiveness of generating an adaptive embedding space for each task to incorporate multiple and richer semantics. Similarly, Zhang et al.~\cite{DBLP:journals/corr/abs-2106-14467} also adopt the adaptive mixture mechanism to combine the generated features which are paired by two conditional variational autoencoders with different modality conditions. The adaptive learning approach is originated from:
\begin{equation}\label{adaptive}
   p_c{'} = \lambda * p_c + (1-\lambda) * g(t_c),
\end{equation}
where $\lambda$ is the adaptive mixture coefficient and $g(t_c)$ is the normalized version of the semantic embeddings lying on the space of the same dimension with the visual prototypes. The above papers have demonstrated that the adaptive method is flexible enough to project the possible modalities onto the same embedding space and provide a way to calculate their corresponding relationships. The adaptive method actually blurs the boundary between them by dictating the modality semantics through the weighted interaction to scaffold the importance of auxiliary modalities on the main modalities. 
\par
In addition to modeling the shifted prototypes explicitly in two modalities and then synthesizing a new joint multimodal prototype, Mu et al.~\cite{DBLP:conf/acl/MuLG20} also attempt to train an embedded network to implicitly add the influence of shifted prototypes to the objective function. They propose language-shaped learning that encourages the original embedding function learned by the visual prototypes to decode the class language descriptions. The objective function $L_M(\theta)$ of the prototypical training is optimized by jointly minimizing the classification loss and the language model loss conditioned on the averaged prototypes:
\begin{equation}\label{joint-loss}
L_M(\theta) = E_{S_e, Q_e} - \sum_{t=1}^{J}log p_{\theta}(y_t|q_t, S_e) - \lambda\sum_{n=1}^{N}\sum_{j=1}^{J_n}log\ g_{\Phi}(w_j|\widetilde{c_n}).
\end{equation}
Different from Eq.~(\ref{loss}), the second item of Eq.~(\ref{joint-loss}) is added as the natural language loss where $\lambda$ controls the weight and $J_n$ is a set of language descriptions for each class. $g_{\Phi}$ is defined as a language model conditioned on $\widetilde{c_n}$ which is the averaged prototype from all the support and query samples of class $n$. There are no apparent representations of the prototypes of the semantic modality, but the shifted effect is reflected in the new embedding function which is learned after introducing the language model generating class semantics conditioned on the visual prototypes.
\par
While most of the prototypical variants are explored for FSL, we cannot ignore the natural use of multimodal information in ZSL where auxiliary modality is essential to solving the problem of no access to visual information in the test phase. To take advantage of textual modalities in the task of zero-shot image classification, Hu et al.~\cite{hu2018correction} do not implement a generative model to adjust the new prototype but use a correction module to update the initial prediction that maps text features to the class centers in the image feature space. The correction module implemented by prototypical networks serves as a meta-learner to adjust the initial prediction to be closer to the target value than the original prediction. After applying the correction, a shifted prototype for each unseen class in the image feature space is obtained and used to classify individual test samples to their nearest cluster centers. \par
The discriminative modality problem could also be exploited to retain projected prototypes in their opposing modality space. Due to the challenge of mode collapse issues in training ZSL tasks, Yu et al.~\cite{DBLP:conf/cvpr/YuJHZ20} propose an episode-based training framework to learn the embedding functions of visual prototype and class semantic prototype, contributing to the accumulated experience on predicting and adapting to unseen image classes. The support set in each episode is used to align semantic interactions between the visual and the class semantic modalities which are processed by two prototype generating networks. The query set is finally evaluated by minimizing multimodal loss $L_M(\theta)$ in both modality spaces $V$ and $T$:\par
\begin{equation}\label{lm-equation}
L_M(\theta) = E_{S_e, Q_e} - \sum_{t=1}^{J}log p_{\theta}^V(y_t|q_t, S_e) - \sum_{t=1}^{J}log p_{\theta}^T(y_t|q_t, S_e).
\end{equation}\par
Arguably, shifted prototypes flexibly represent, select, and combine multiple modalities (usually two). It is particularly fundamental for the imbalanced semantic distribution of modalities in the low-data context. Although it is not complicated to apply Prototypical Networks, cautions are indispensable for encoding different modalities into the objective function. The choice of the weighting factor and the mapping network structure for the alignment will affect the modality knowledge that the meta-learning framework can eventually transfer. Another concern is that good prototypes are built on the assumption of the existence of a single prototypical representation for each class, which may not apply to some skewed data. \par
\subsection{Fused-Modal Matching}
\subsubsection{Overview}
Vinyals et al.~\cite{DBLP:conf/nips/VinyalsBLKW16} propose Matching Networks to predict the one-hot encoded labels of the query set by adding attention and memory mechanisms into one-shot training. It explicitly implements the set-to-set paradigm to learn from the given support set to minimize a loss over a batch. The predictions for each new query sample is defined by the parametric neural network $P(\hat{y}|\hat{x}, S)$, which can be formulated as a linear combination of the labels in the support set $S$ with the help of the attentional kernel $a(\hat{x}, x_i)$: \begin{equation}\label{matching-network}
P(\hat{y}|\hat{x}, S) = \sum_{i \in S} a(\hat{x}, x_i)y_i.
\end{equation}
\par
With results mapping from unimodal representation to the multimodal space, computed attention is modified through the normalized similarity comparison in which the query sample and each sample in the support set are supposed to incorporate multiple modalities into their embeddings. \par
\subsubsection{Methods}
The idea of implicitly representing modalities in Matching Networks inspires researchers to wrap multimodal information into the kernel function. Yao-Hung and Ruslan~\cite{DBLP:journals/corr/abs-1710-08347} introduce the statistical dependence measurement of fusing auxiliary modalities with the original image data, and then combine the attention mechanism of Matching Networks to learn the probability distribution of ``lots-of-examples'' on the labels of ``one-example'' classes. They express auxiliary modalities as prior knowledge representing relationships between categories, including supervised human-crafted features and unsupervised word embeddings from language models. During the modality fusion stage, they adopt HSIC~\cite{gretton2005measuring} to learn the kernels that can integrate different types of modalities and transfer knowledge across classes. The learned label affinity kernels $a_r(y, y'_{i})$ will then be used for the prediction of label probability distribution $P(y'|y;R,R')$ on the label space for ``one-example'' classes:
\begin{equation}\label{kernel-match}
P(y'|y;R,R')=\sum_{i \in S'}a_r(y, y'_{i})y'_{i},
\end{equation}
where $a_r(\cdot,\cdot)$ is the attentional kernel mapping from the modalities $R$ of ``lots-of-examples'' classes to the modalities $R'$ of ``one-example'' classes. $S$ stands for the support set with data-label pairs $\{x_i, y_i\}$ and $S'$ denotes a different support set with ``one-example'' data $\{x^{'}_i, y^{'}_i\}$.
The computation is similar to the label prediction for each query sample in Matching Networks (See Eq.~(\ref{matching-network})), with the difference of whether the kernels contain multimodal information. Finally, they combine the regression model and the non-parametric attention model in the loss function, and construct a meta-learning strategy~\cite{DBLP:conf/nips/VinyalsBLKW16} to achieve the adaptation from ``lots-of-examples'' classes to ``one-example'' classes.\par
The end-to-end architecture of Matching Networks is convenient to produce predictions for the unseen classes while new changes to the networks are bypassed. Besides, the attention kernel is inclusively adapted to multimodal information in tandem with specifying the final classifier. However, due to the weak comprehensibility of the kernel function, it is difficult to understand how different modalities work together in conveying semantics.

\subsection{Related-Modal Operator}
\subsubsection{Overview}
While the above networks focus on the embeddings that can be compared by a pre-defined metric or a linear classifier, Sung et al.~\cite{DBLP:conf/cvpr/SungYZXTH18} propose Relation Networks that learn the similarity comparison by a learnable non-linear operator and produces a similarity score between 0 and 1 within a relation module. The objective function is to minimize the sum of mean square losses $L$ of the relation scores $r_{i,j}$ to their ground truths contained in the support set and query set: 
\begin{equation}\label{relation-loss}
    L = \sum_{i\in S_e}\sum_{j\in Q_e}(r_{i,j}-\mathbb{I}(y_i == y_j))^2.
\end{equation}
\par
When the features are multimodal, instead of only feeding the unimodal query and the support samples to the relation module after being preprocessed by the same embedding module, the samples are first processed by heterogeneous embedding modules based on their modality pattern and then concatenated together as a whole feature map for the input of the relation module. 
\subsubsection{Methods}
Sung et al.~\cite{DBLP:conf/cvpr/SungYZXTH18} expand the application of Relation Networks from FSL to ZSL/GZSL, and train two auxiliary parameterized networks for visual and semantic modalities on the embedding modules. They perform episode-based learning to exploit the support set and the query set in the network optimization process of meta-parameters. The relation score for each query sample is obtained by applying the relation net on the concatenated multimodal feature map. However, such Relation Networks can be considered simple binary decision-making regression models, which directly measure the difference between the ground truth and the predicted label, resulting in the limited ability to predict labels in an end-to-end framework.\par

\begin{figure*}[ht]
     \centering
     \begin{subfigure}[b]{0.44\textwidth}
         \centering
         \includegraphics[width=\textwidth]{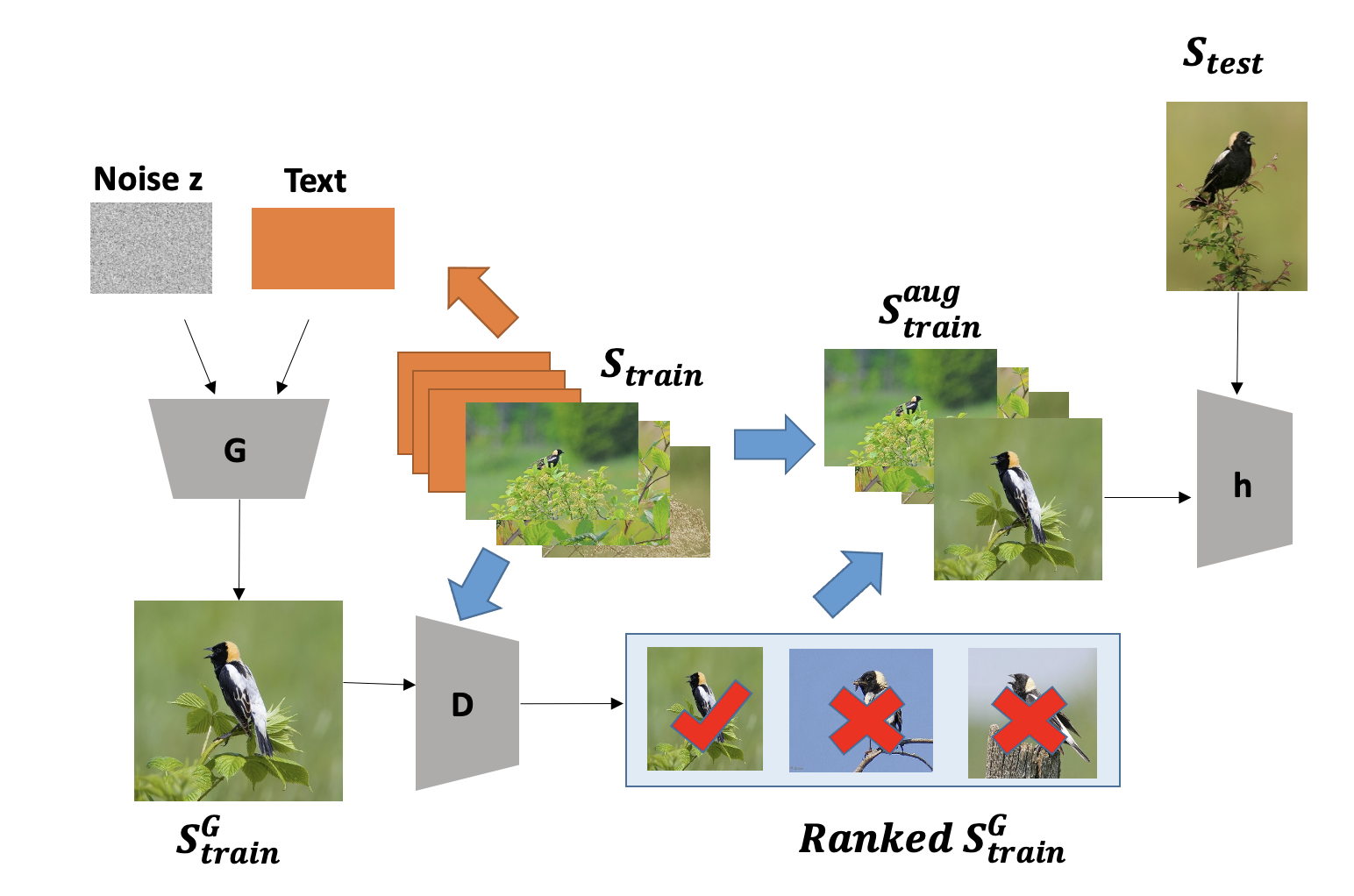}\vspace{2ex}     
     \end{subfigure}
     \hspace{10pt}
     \begin{subfigure}[b]{0.44\textwidth}
         \includegraphics[width=\textwidth]{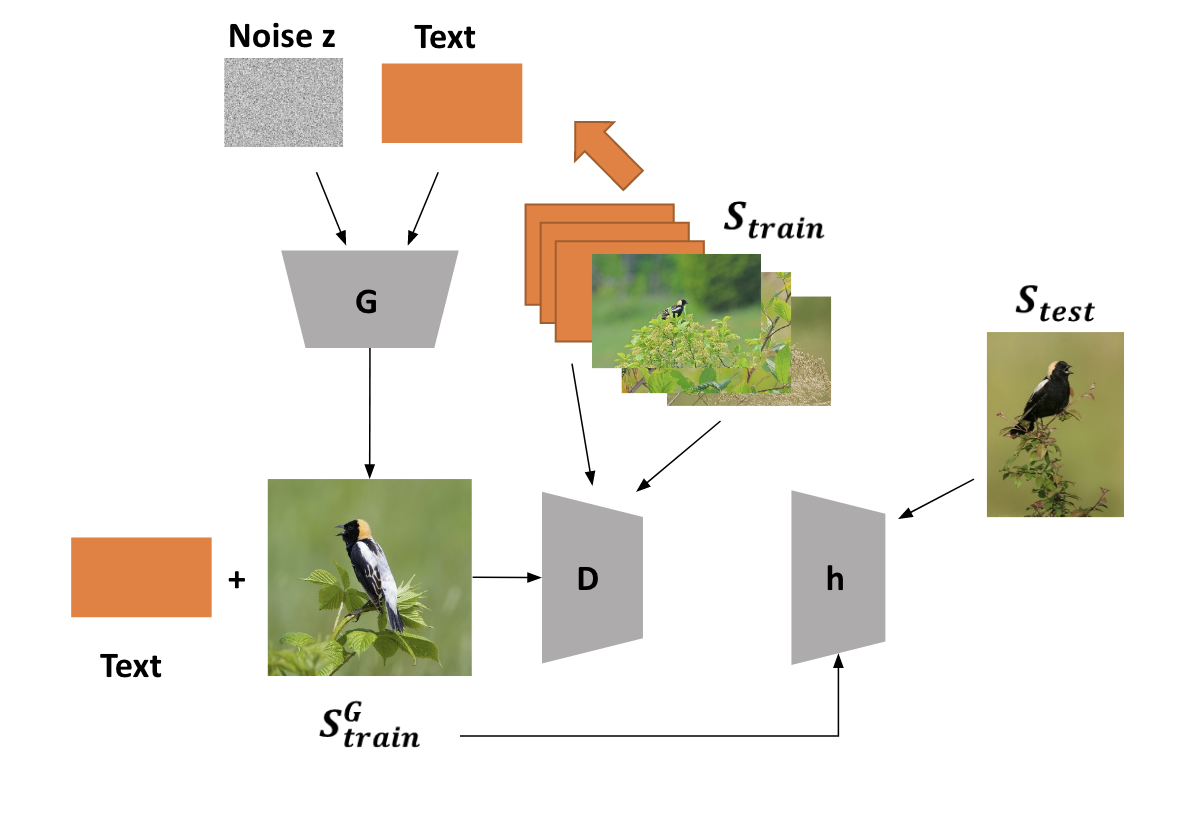}
     \end{subfigure}
        \caption{Network structures for cross-modal augmentation. \textbf{Left}: Adapted from~\cite{DBLP:conf/cvpr/WangGHH18,DBLP:conf/wacv/PahdeOJKN19}. Meta-learning with hallucination. Given an initial training set $S_{train}$ with two modalities, we combine a ranked $S_{train}^G$ generated by the self-paced tcGAN with $S_{train}$ to create an augmented training set $S_{train}^{aug}$. $S_{train}^G$ is obtained by conditioning on the textual modality and noise vectors $z$. The generative network is trained jointly with the classifier $h$. \textbf{Right}: Adapted from~\cite{DBLP:conf/aaai/VermaBR20}. For each task, the support set is used by the meta-learners $G$ and $D$ in the inner loop. The classifier $h$ takes the generated sample from $G$ and classifies it into the original class.}
        \label{hallucinator}
\end{figure*}

To further exploit the discriminating power of the metric loss for GZSL problems, CPDN~\cite{DBLP:journals/spl/HuangLH20} formulates the relation model as a cross-modal metric learning problem. Similarly, the embedding module is still retained as the feature encoder to extract visual features. By conditioning on the textual attributes of unseen classes, a generative network of generating the visual prototype $p_j$ for each class is constructed. Then, a new discriminative relation network is presented for learning a non-linear metric space to measure similarities between the image samples and the generated class-prototypes. The learning process adopts the standard episode-based meta-learning framework~\cite{DBLP:conf/icml/FinnAL17}. One major modification is that two novel metric losses, prototype-sample metric loss $L_{PS}$ and class-prototype scatter loss $L_{CS}$ are added to the original label prediction loss $L_{LP}$. The new loss function $L_{new}$~\cite{DBLP:journals/spl/HuangLH20} is shown in:
\begin{equation}\label{modified-relation}
    L_{new} = L_{LP} + \lambda L_{PS} + \beta L_{CS},
\end{equation}
where 
\begin{equation}
    L_{LP} = \sum_{i=1}^n\sum_{j=1}^c(r_{i,j}-\mathbb{I}(x_i, p_j))^2,
\end{equation}
\begin{equation}
    L_{PS} = 2\sum_{i=1}^n\sum_{j=1}^c(\frac{1}{2}-\mathbb{I}(x_i, p_j))r_{i,j},
\end{equation}
\begin{equation}
    L_{CS} = \sum_{i=1}^n\sum_{j=1}^c(1-\mathbb{I}(p_i, p_j))\widetilde{r}_{i,j}.
\end{equation}
$r_{i,j}$ is the relation score of the $i$-th sample and the $j$-th class-prototype. $\widetilde{r}_{i,j}$ is the relation score of the $i$-th and $j$-th prototypes.\par
Due to the scalability of the embedding module, another modality can be freely merged with the main modality, thereby achieving the realization of the multimodality-based Relation Networks. This also explains the fluency of migrating the network structures from FSL to ZSL/GZSL. Nonetheless, the risk is to assume that one modality has full predictive power for another. The above two methods have not realized the simultaneous modeling of two modalities that describe the same object, let alone extended to more modalities. Research gaps in the multimodality-based FSL still exist. In addition, we may infer that since the Relation Networks can be applied as a simple binary decision-maker, it is desirable to implement them as the final step to assist prediction in other models. For example, the implemented solution~\cite{DBLP:journals/spl/HuangLH20} includes the metric loss calculated by various prototypes to the original MSE loss.

\section{Learn the Generation}\label{section-data}
The distribution of the data itself can be modeled as learnable information, which is common in traditional statistical learning and ML. With the continuous emergence of neural networks, powerful networks provide more complex modeling methods to learn data distribution, especially for FSL/ZSL. The meta-learning framework also tries to enhance the generalization ability by learning more meta-knowledge about the data. The most common practice is introducing a generative neural network to learn its ability to generate data and use it in new tasks to enrich a small number of samples. We thus review relevant studies utilizing multimodal data to model flexible generation processes combined with the meta-learning framework.

\subsection{Cross-Modal Augmentation}
\subsubsection{Overview}
Data augmentation is another intuitive method for training small amounts of labeled data. The common techniques are often applied to generate extra samples in the image domain simply by a set of transformations, such as flipping, rotating, adding Gaussian perturbations and deformation~\cite{DBLP:conf/cvpr/ChenFW00H19,DBLP:conf/aaai/ChenFCJ19,DBLP:journals/tip/ChenFZJXS19}. However, these transformations only modify the geometric structure of the image and preserve some original properties like parallelism of lines or ratios of distances. Due to the lack of prior knowledge about data distributions, it is difficult to generate more complex new samples. \par
In order to prevent overfitting in supervised learning, it is important to increase the number of data samples. As an analogy, a meta-learning algorithm's performance and generalization abilities mainly depend on the number and diversity of tasks. However, it is not always feasible to obtain sufficient meta-training tasks. Therefore, similar to data augmentation, it is intuitive to use task augmentation to generate more meta-training tasks. Liu et al.~\cite{liu2020task} simply rotate the images to increase the number of classes. Yao et al.~\cite{yao2021improving} linearly combine both features and labels of samples within the same meta-training task. Yao et al.~\cite{yao2021meta} generate new tasks by interpolating features and labels of a pair of samples that are randomly sampled from different meta-training tasks. Task augmentation transforms and combines features and labels to generate more tasks, but it is not guaranteed that the generated meta-training tasks are complex and diversified due to the lack of prior task distribution. \par
Recently, more advanced data augmentation methods have been studied to learn new concepts from few labeled samples. Hariharan and Girshick~\cite{DBLP:conf/iccv/HariharanG17} consider the intra-class variations and learn the analogical relations between paired samples. VI-Net~\cite{DBLP:conf/iccv/ZhangZNXY19} constructs a sampling space where the distribution is generated from the sharing class-specific pool of support sets and query sets. Such a variational Bayesian network can then be combined with any classifier-free mechanism to predict the novel classes. DAGAN~\cite{DBLP:journals/corr/abs-1711-04340} is built upon WGAN~\cite{DBLP:conf/icml/ArjovskyCB17} to augment the support sets in the one-shot learning and sample from the space that is related to the best manifolds, facilitating the classification performance of matching networks. However, generic generative methods mostly transform the FSL into standard supervised learning after generating data distributions or enough samples. They usually start with base classes and then migrate the learned data pattern to generate new samples of the novel classes. One major weakness is that the generative model is trained separately from the classifier so that no explicit prior knowledge can be quickly adapted to novel classes.\par
The recently emerging meta-learning framework has spawned relevant work on the meta-based generative methods to learn an optimal augmentation strategy that can be trained within a few steps given unseen classes. The main idea is to wrap up the data augmentation strategy in optimization steps of the inner meta-learning process~\cite{DBLP:journals/corr/abs-2004-05439}, with the augmentation parameterized and learned by the outer optimization in the episode training of meta-representation. MetaGAN~\cite{DBLP:conf/nips/ZhangCGBS18} explores the impact of generating fake data on the decision boundary and provides ways to integrate the model with MAML or Relation Networks. Wang et al.~\cite{DBLP:conf/cvpr/WangGHH18} rely on a simple multi-layer perceptron as a hallucinator learned on the base classes to synthesize new data for unseen classes. However, the generalization ability could still be limited due to the lack of additional modalities that are helpful for the model to identify highly discriminative features across modalities~\cite{DBLP:journals/corr/abs-1806-05147} in the low data regime. Cross-modal data is considered an input that can effectively assist in data augmentation. By mining the relationship between the two modalities, the modality with insufficient information can be enriched.

\begin{table*}[h]
    \centering
    \caption{Summary of multimodality-based meta-learning methods. ``V'', ``T'', ``S'', ``Te'' and ``A'' indicate visual modalities, textual modalities, spatio modalities, temporal modalities and audio modalities, respectively. ``M'' and ``F'' indicate missing and full modalities, respectively.}
    \resizebox{\textwidth}{!}{
    \begin{tabular}{lllll}
    \hline
    Reference & Type & Modality & Dataset & Application \\
    \hline
    Ma et al.~\cite{DBLP:conf/aaai/MaRZTWP21} & Optimization & V(F), T(M)/A(M) & MM-IMDb, CMU-MOSI, avMNIST & Image/multi-genre/binary sentiment classification \\
    Yao et al.~\cite{DBLP:conf/www/YaoLWTL19} & Optimization & S(F), Te(F) & City mobility/water quality datasets & Traffic volume/water quality prediction\\
    Yan et al.~\cite{DBLP:conf/iros/YanLSY20} & Optimization & V(F), A(F) & AI2-THOR & Indoor navigation\\
    Verma et al.~\cite{DBLP:journals/corr/abs-2102-11856} & Optimization & V(F), T(F) & CUB-200, aPY, AWA1, AWA2, SUN & Image classification\\
    Liu et al.~\cite{DBLP:conf/aaai/LiuLY0L21} & Optimization & V(F), T(F)&CUB-200, aPY, AWA1, AWA2 & Image classification\\
    Ma et al.~\cite{ma2021model} & Optimization & V(F), T(F) & USer-InstaPIC/-YFCC & Cross-modal retrieval\\
    Chen and Zhang~\cite{DBLP:journals/corr/abs-2105-07889}   & Optimization & V(F), T(F) & CUB-200, ModelNet40& Image classification\\
    Liang et al.~\cite{DBLP:journals/corr/abs-2012-02813} & Optimization & V(F), T(F); V(F), A(F); T(F), A(F) & Yummly-28K,  CIFAR-10/-100, ESC-50, Wilderness & Image/speech classification\\
    Ge and Xiaoyang~\cite{DBLP:conf/iccvw/SongT19} & Embedding & V(F), T(F) & Wiki, MIRFLICKR, NUS-WIDE & Cross-modal retrieval\\
    Liu and Zhang~\cite{DBLP:journals/corr/abs-2009-07879} & Embedding & V(F), A(F) & COIL-100 & Visual learning\\
    Eloff et al.~\cite{DBLP:conf/icassp/EloffEK19} &  Embedding & V(F), A(F) & MNIST, TIDigits & Cross-modal matching \\
    Nortje and Kamper~\cite{DBLP:journals/corr/abs-2012-05680} & Embedding & V(F), A(F) & MNIST, TIDigits, Buckeye & Cross-modal matching\\
    Wan et al.~\cite{DBLP:conf/aaai/WanZDHYP21}  & Embedding & V(F), T(F) & TV series, e-book pieces & Social relation extraction \\
    Pahde et al.~\cite{DBLP:conf/wacv/PahdePKN21} & Embedding & V(F), T(F) & CUB-200, Oxford-102 & Image classification\\
    Xing et al.~\cite{DBLP:conf/nips/XingROP19}  &  Embedding & V(F), T(F) & CUB-200, miniImageNet, tieredImageNet& Image classification\\
    Schwartz et al.~\cite{DBLP:journals/corr/abs-1906-01905} & Embedding & V(F), T(F) & CUB-200, miniImageNet & Image classification\\
    Zhang et al.~\cite{DBLP:journals/corr/abs-2106-14467} & Embedding & V(F), T(M) & CUB-200, miniImageNet, CIFAR-FS  & Image classification\\
    Mu et al.~\cite{DBLP:conf/acl/MuLG20} & Embedding & V(F), T(F) & CUB-200, ShapeWorld & Image classification\\
    Hu et al.~\cite{hu2018correction} & Embedding & V(F), T(F) & CUB-200, NAB & Image classification\\
    Yu et al.~\cite{DBLP:conf/cvpr/YuJHZ20}  &  Embedding & V(F), T(F) & CUB-200, AWA1, AWA2, Oxford-102 & Image classification \\
    Yao-Hung and Ruslan~\cite{DBLP:journals/corr/abs-1710-08347} &  Embedding & V(F), T(F) & CUB-200, AWA1 & Image classification \\
    Sung et al. \cite{DBLP:conf/cvpr/SungYZXTH18} & Embedding & V(F), T(F)& CUB-200, miniImageNet, AWA1, AWA2, Omniglot & Image classification\\
    Huang et al.~\cite{DBLP:journals/spl/HuangLH20} &  Embedding & V(F), T(F) &  CUB-200, AWA1, AWA2 & Image classification\\
    Pahde et al.~\cite{DBLP:journals/corr/abs-1806-05147} & Generation & V(F), T(F)& CUB-200 &Image classification\\
    Verma et al.~\cite{DBLP:conf/aaai/VermaBR20} &  Generation & V(F), T(F) & CUB-200, aPY, AWA1, AWA2, SUN & Image classification\\
    \hline
    \end{tabular}
    }
    \label{overview-analysis}
\end{table*}
\subsubsection{Methods}
Pahde et al.~\cite{DBLP:journals/corr/abs-1806-05147} extend the model from Wang et al.~\cite{DBLP:conf/cvpr/WangGHH18} in a multimodal and progressive manner by hallucinating images conditioned on fine-grained textual modality within the few-shot scenario. They propose a text-conditional GAN (tcGAN) to learn the mapping from the textual space to visual space by a self-paced based generative model~\cite{DBLP:conf/wacv/PahdeOJKN19}, which ensures that only the top-ranked generated image samples are selected and aggregated with the original novel samples. As shown in Figure \ref{hallucinator} (Left), the hallucinator is parameterized as $G(x_i^{text}, z_i; w_G)$ and the discriminator is parameterized as $D(x_i^{image}; w_D)$, both connected to the inner classifier $h(x, S_{train}^{aug}; w_C)$ through the generated $S_{train}^{aug}$. The parameters $w_G$ and $w_D$ from the generative network and $w_C$ from the few-shot classifier are updated directly by a joint back-propagation in the meta-training stage to augment data that are useful for discriminating classes. \par
The generative model requires optimized generator and discriminator parameters, which are naturally suitable for the inner-optimization of the MAML training framework. Verma et al.~\cite{DBLP:conf/aaai/VermaBR20} identify the key challenge of using generative models to train ZSL/GZSL by leveraging auxiliary modalities. They propose a meta-based generative framework that integrates MAML with a generative model conditioned on class attributes for unseen samples generation. The key difference with the standard meta-learning is to mimic the ZSL behavior by setting disjoint meta-train and meta-validation partitions for each task. The meta-learning protocol modifies the standard adversarial generation process to provide an efficient discriminator and generator by enhancing their learning capability. \par
After the data augmentation strategy is parameterized as the inner optimizer to participate in the training, it still relies on the optimization-based meta-learning framework, except that the inner learner no longer pays attention to the learning of multimodal relations. More diversified data generators beyond conditional GANs are expected to be introduced in the future and combined with metric-based meta-learners. 

\section{Summary of Methods}\label{section-overview}
This section provides a summary of recent works about multimodality-based meta-learning methods. The attributes of each method are concluded in Table \ref{overview-analysis}. Specifically, we highlight the following attributes and discuss some common applications.
\par
\begin{itemize}
    \item \textbf{Reference:} It denotes the individual papers that propose the method. Table \ref{source-code} lists the open-source libraries for each reference if available.
    \item \textbf{Type:} It indicates the type of meta-knowledge learned by the method, including learning the optimization, learning the embedding, and learning the generation.
    \item \textbf{Modality:} It summarizes the modalities and their patterns employed in the method. Missing modalities apply to those with partial data missing, and full modalities mean that the involved modalities are not missing in the datasets. Most studies address the two most common ones, visual and textual modalities, to analyze image-related applications.
    \item \textbf{Dataset:} It introduces datasets that are used to evaluate the method. Many works employ CUB-200, AWA1, and AWA2 as the major datasets to derive multimodal information in the application of image classification. Table \ref{overview-datasets} lists open-source datasets used in papers if available.
    \item \textbf{Application:} It gives the application fields of the method, most of which are summarized based on the evaluation experiments of the paper. A large number of studies address image classification by using auxiliary textual modality. Other important areas are classification-based (e.g., sentiment, speech), prediction-based (e.g., traffic volume, water quality), and cross-modal retrieval/matching.
\end{itemize}

\begin{table}[h]
    \scriptsize
    \centering
    \caption{List of open-source libraries for multimodality-based meta-learning methods.}
    \begin{tabularx}{\columnwidth}{lX}
    \hline
    Reference & Link \\
    \hline
    Ma et al. \cite{DBLP:conf/aaai/MaRZTWP21} & \url{https://github.com/mengmenm/SMIL} \\
    Yao et al. \cite{DBLP:conf/www/YaoLWTL19} & \url{https://github.com/huaxiuyao/MetaST} \\
    Liang et al. \cite{DBLP:journals/corr/abs-2012-02813} & \url{https://github.com/peter-yh-wu/xmodal} \\
    Eloff et al. \cite{DBLP:conf/icassp/EloffEK19} & \url{https://github.com/rpeloff/multimodal\_one\_shot\_learning}\\
    Nortje and Kamper \cite{DBLP:journals/corr/abs-2012-05680} & \url{https://github.com/LeanneNortje/DirectMultimodalFew-ShotLearning} \\
    Wan et al. \cite{DBLP:conf/aaai/WanZDHYP21} & \url{https://github.com/sysulic/FL-MSRE} \\
    Xing et al. \cite{DBLP:conf/nips/XingROP19} & \url{https://github.com/ElementAI/am3} \\
    Mu et al. \cite{DBLP:conf/acl/MuLG20} & \url{https://github.com/jayelm/lsl} \\
    Yu et al. \cite{DBLP:conf/cvpr/YuJHZ20} & \url{https://github.com/yunlongyu/EPGN} \\
    Sung et al. \cite{DBLP:conf/cvpr/SungYZXTH18}  & \url{https://github.com/floodsung/LearningToCompare\_FSL}\\
    Verma et al. \cite{DBLP:conf/aaai/VermaBR20}  &  \url{https://github.com/vkverma01/meta-gzsl}\\
    \hline
    \end{tabularx}
    \label{source-code}
\end{table}  

\begin{table}[ht]
    \scriptsize
    \centering
    \caption{List of open-source datasets used for experimental evaluations of multimodality-based meta-learning methods.}
    \begin{tabularx}{\columnwidth}{lX}
    \hline
    Dataset & Link \\
    \hline
     MM-IMDb & \url{http://lisi1.unal.edu.co/mmimdb/}\\
     CMU-MOSI & \url{http://multicomp.cs.cmu.edu/resources/cmu-mosei-dataset/}\\
     avMNIST & \url{https://arxiv.org/pdf/1808.07275.pdf} \\
     AI2-THOR & \url{https://github.com/allenai/ai2thor} \\
     CUB-200 & \url{http://www.vision.caltech.edu/visipedia/CUB-200.html}\\
     aPY & \url{https://vision.cs.uiuc.edu/attributes/}\\
     AWA1 & \url{https://cvml.ist.ac.at/AwA/}\\
     AWA2 & \url{https://cvml.ist.ac.at/AwA2/}\\
     SUN & \url{https://groups.csail.mit.edu/vision/SUN/hierarchy.html}\\
     USer-InstaPIC/-YFCC & \url{https://ieeexplore.ieee.org/document/9465686/}\\
     ModelNet40 & \url{https://modelnet.cs.princeton.edu/} \\
     Yummly-28K & \url{http://www.lherranz.org/datasets/}\\
     CIFAR-10/-100 & \url{https://www.cs.toronto.edu/~kriz/cifar.html}\\
     ESC-50 & \url{https://github.com/karolpiczak/ESC-50}\\
     Wilderness & \url{https://github.com/festvox/datasets-CMU\_Wilderness}\\
     Wiki & \url{https://dl.acm.org/doi/10.1145/1873951.1873987}\\
     MIRFLICKR & \url{https://press.liacs.nl/mirflickr/}\\
     NUS-WIDE & \url{https://lms.comp.nus.edu.sg/wp-content/uploads/2019/research/nuswide/NUS-WIDE.html}\\
     COIL-100 & \url{https://www.cs.columbia.edu/CAVE/software/softlib/coil-100.php}\\
     MNIST & \url{http://yann.lecun.com/exdb/mnist/}\\
     TIDigits & \url{https://catalog.ldc.upenn.edu/LDC93S10}\\
     Buckeye & \url{https://buckeyecorpus.osu.edu/}\\
     Oxford-102 & \url{https://www.robots.ox.ac.uk/~vgg/data/flowers/102/}\\
     miniImageNet & \url{https://github.com/yaoyao-liu/mini-imagenet-tools}\\ 
     tieredImageNet & \url{https://github.com/yaoyao-liu/tiered-imagenet-tools}\\
     ShapeWorld & \url{https://github.com/AlexKuhnle/ShapeWorld}\\
     NAB & \url{https://dl.allaboutbirds.org/nabirds}\\
     Omniglot & \url{https://www.tensorflow.org/datasets/catalog/omniglot}\\ 
    \hline
    \end{tabularx}
   
    \label{overview-datasets}
\end{table}
Furthermore, we summarize the advantages and disadvantages for each method in Table \ref{comparison-table}. The advantage that commonly appears in papers involves the flexible architecture designed for auxiliary modalities or specific tasks. The integration of feature spaces has been widely explored, though requirements for transferring models to multiple modalities or more generalized task distributions are still needed.
\subsection{Common Applications}
\textbf{Image classification}\label{image-cl}: Image classification is one of the most common applications in the field of supervised meta-learning, where predictions are made on unknown images based on a model trained on a set of tasks. With the addition of multimodalities in FSL/ZSL, the model can better utilize textual or audio information to understand the visual modalities, such as alignment, matching, and fusion. \par
\textbf{Multi-genre classification}\label{mul-gen-cl}: Multi-genre classification is a specific type of multi-label classification, which assumes that each sample can be assigned to more than one category. Genres are often accompanied by multiple modalities such as textual attributes, meta-attributes, movie images, etc. The introduction of meta-learning helps to better identify the effects of different modalities on different categories.\par

\begin{table*}[t]
    \scriptsize
    \centering
    \caption{Comparison of multimodality-based meta-learning methods.}
    \begin{tabularx}{\textwidth}{llXX}
    \hline
    Method & Reference & Advantages & Disadvantages \\
    \hline 
    \multirow{4}{*}{Parameterized-modal initialization} & \cite{DBLP:conf/aaai/MaRZTWP21} & Efficient and flexible with the integration of feature reconstruction and training process & Sensitive to prior selection in the Bayesian framework \\
    & \cite{DBLP:conf/www/YaoLWTL19} & Transfers sequential knowledge from multiple sources & Strong dependence on the network structure\\
    & \cite{DBLP:conf/iros/YanLSY20} & Adapts multimodal signals and loss functions to the RL setting & Requires the RL objectives to be identifiable and learnable\\
    & \cite{DBLP:journals/corr/abs-2102-11856} & Handles streaming data for the ZSL & Difficult to scale to multiple modalities\\
    & \cite{DBLP:conf/aaai/LiuLY0L21} & Robust to class-aware task distributions & Complex learning process for synthesized features \\
    & \cite{ma2021model} & Generalizes well to unknown user domains without updating, scalability for larger domains & Custom weighted gradient aggregation strategy needs to be considered\\
    & \cite{DBLP:conf/cikm/ChenZ21} & Extends MAML to heterogeneous task distributions & Dependence on the feature aggregation strategies\\
    & \cite{DBLP:journals/corr/abs-2012-02813} & Ensures alignment between source and target modalities, fits strongly and weakly paired data & Lacks discussions on choices of the alignment loss  \\
    \hline 
    Unified-modal optimizer & \cite{DBLP:conf/iccvw/SongT19} & Sequentially combines multiple modalities to a unified space & Less effective without a proper updated rule \\
    \hline
    \multirow{2}{*}{Paired-modal networks} & \cite{DBLP:conf/cvpr/HadsellCL06}, \cite{DBLP:conf/icassp/EloffEK19} & Add loss variants to help with the data deficiencies & Cost-intensive training with paired samples grows as the number of modalities increases\\
    & \cite{DBLP:journals/corr/abs-2012-05680} & Realizes cross-modal shared space &  Pair mining process is time consuming\\
    \hline 
    \multirow{5}{*}{Joint-modal prototypes} & \cite{DBLP:conf/aaai/WanZDHYP21} & Easy to extend to multiple modalities & Lacks theoretical assumptions to support the empirical results \\
    & \cite{DBLP:conf/wacv/PahdePKN21} & Leverages the auxiliary modalities to enrich the main modalities & One-side conditioned relationship between modalities makes it hard to expand to multiple modalities\\
    & \cite{DBLP:conf/nips/XingROP19}, \cite{DBLP:journals/corr/abs-1906-01905}, \cite{DBLP:journals/corr/abs-2106-14467} & Adaptively combine feature across modalities rather than alignment & Limited to two modalities \\
    & \cite{DBLP:conf/acl/MuLG20} & No explicit joint-modeling of prototypes is needed & Distinct properties about different modalities are ignored \\
    & \cite{hu2018correction}, \cite{DBLP:conf/cvpr/YuJHZ20} & Avoid the instability and expensive training of parameters of generative methods for ZSL/GZSL &  Rely on the design of mapping networks \\
    \hline
    Fused-modal matching & \cite{DBLP:journals/corr/abs-1710-08347} & Fast training as an end-to-end architecture & Weak comprehensibility of the kernel-based representation \\
    \hline
    Related-modal operator & \cite{DBLP:conf/cvpr/SungYZXTH18}, \cite{DBLP:journals/spl/HuangLH20} & Compatible with multiple low-data applications by learnable non-linear operators & Rigid in combining modalities \\
    \hline
    Cross-modal augmentation & \cite{DBLP:journals/corr/abs-1806-05147}, \cite{DBLP:conf/aaai/VermaBR20} & Set data augmentation strategies as the optimizee & Heavily depends on the conditional generative model\\
    \hline
    \end{tabularx}
    \label{comparison-table}
\end{table*}

\textbf{Cross-modal retrieval}\label{cross-modal-ret}: Cross-modal retrieval is concerned with mutual retrieval across modalities and focuses on getting semantically similar instances in another modality (e.g., text) by searching in one modality (e.g., image). It is often necessary to find a common representation space to compare samples from different modalities directly. \par
\textbf{Speech classification}\label{audio-cls}:
Speech classification is targeted at automatically classifying speech or audio clips. In addition to leveraging only acoustic features, cross-modal speech classification also introduces visual or textual information.\par
\textbf{Visual learning}\label{visual-l}: Visual learning is a type of learning that takes advantage of graphic information, where learners learn the material by seeing graphics, colors, and other visual information. Traditional visual learning tends to learn relevant concepts from a single pattern of data, neglecting the cross-modal learning abilities of humans. Multimodal speech-vision learning enables phonetic name features and visual object features to be combined with unsupervised meta-learning strategies for parsing visual information acquired at different times and places.\par
\textbf{Cross-modal matching}\label{cross-modal-mat}: Cross-modal matching focuses on recognizing the same objects presented in two different modalities, usually by matching a set in one modality with a matching set in another modality. For example, matching speech segments with image information, matching image information with textual information, etc.\par
\section{Discussions and Conclusions}\label{section-dis}
We have reviewed the multimodality-based meta-learning methods and classified the existing algorithms into learning the optimization, learning the embedding, and learning the generation based on the learned meta-knowledge. From the publication year of the papers reviewed in this article, it can be seen that in the past two years, meta-learning methods based on multimodal datasets have gradually attracted the community's attention. Their main entry point is to use FSL as the research objective, and then use the characteristics of multimodality to make up for the training inefficiencies caused by the primary modalities in the few samples. At the same time, a specific meta-learning algorithm is used as the pivot and then the multimodal patterns are adjusted to adapt to the input or training process of the algorithm. As a result, the ability of meta-learning algorithms to quickly adapt to new tasks helps improve the overall performance.\par
The first meta-learning strategy we inspect is the optimization of model parameters (e.g., initialization, hyperparameters of the optimizer), which can help train new tasks within a few updates. Since the variant networks based on MAML or extended from MAML have achieved great success on the unimodal datasets, most researchers focus on modifying the MAML-like structure accompanied by the multimodal input and training process. We find that modifying within-task alignment mainly incorporates the parameterized multimodal model as the inner learner and still learns the initialization used for unseen tasks. A more popular way for cross-task alignment, in contrast, prefers to integrate the task-dependent parameters into the meta-learning structure and output the partially shared meta-knowledge including these parameters. Notably, the advantage of optimization-based meta-learning methods is that the variants are for the training framework without affecting how the multimodal models are encoded. True multimodalities can be guaranteed instead of being limited to common or paired modalities. \par
The lack of multimodal training datasets with few samples is often regarded as an obstacle to applying multimodality-based meta-learning algorithms. However, this notion is only partially correct. It can be seen from Table \ref{overview-analysis} that the application fields are extensive, and the datasets are not limited to image classification. As the applications of meta-learning in real environments are introduced with more public datasets available, the flexibility of the optimization-based framework is expected to play a more important role in adapting to new changes of modality patterns. \par
From the number of article distributions (See Table \ref{overview-analysis}), we have seen that the metric-based meta-learning algorithms have become the preferred approach for interpreting multimodal data. In fact, metric-based algorithms provide exactly what can help multiple modalities map to the same embedding space, thus implicitly solving the problem originated from traditional multimodal ML and providing convenience to compare distances using the metric. Specifically, after reviewing relevant papers using Prototypical Networks to capture changes brought by multimodal prototypes, one would expect to be able to summarize the perfect prototype fusion method. Although different methods have proved that their performance exceeds that of unimodal prototype applications, a striking conclusion we can draw is that there is no exact solution to uniformly solve the problem of modifying multimodal prototypes in different applications. For example, in the adaptive modality mixture mechanism (AM3), many researchers~\cite{DBLP:conf/nips/XingROP19,DBLP:journals/corr/abs-1906-01905,DBLP:journals/corr/abs-2106-14467} use the same prototypical networks and modality types, but the results are very different. A key aspect that is often overlooked is that the discriminative features that different modalities can provide are very different in various scenarios. Moreover, it also depends on the dataset itself. Therefore, the determination of which modality is more important is not a simple procedure. The benefits that multimodal prototypes can provide are not only to expand the possibility of more modality combinations in the embedding space, but may also be affected by the previous encoding models of multimodal data. However, these factors hardly appear in the current works.\par
Researchers who have achieved good performance usually distinguish aspects other than the structure of meta-learning algorithms, such as new data distributions. This philosophy is embodied in learning data generation. Although Pahde et al.~\cite{DBLP:conf/wacv/PahdeOJKN19} and Verma et al.~\cite{DBLP:conf/aaai/VermaBR20} have verified the improved performance of the results brought by data augmentation strategies, there are no further discussions about whether the design of a specific meta-learning architecture containing multimodal tasks can be combined with data augmentation methods to obtain better results. Furthermore, we have noticed the potential of incorporating the meta-based ZSL/GZSL methods into our consideration. These methods divide training classes into support sets and query sets, enabling the augmentation process to learn a robust mapping when there are only auxiliary modalities or few samples. The prejudice against the seen classes during training is thus alleviated. \par
In conclusion, the academic community has paid much attention to specific applications of multimodality-based meta-learning models. New frameworks are applied to multimodal datasets and compared with the baselines to illustrate their advantages. However,  theoretical analyses about mathematical assumptions, generalized performance, complexity, and convergence of models remain blank. 

\section{Future Directions}\label{section-fut}
This section provides some insights about future directions based on the current challenges and trends. We believe that the summary provided in this survey can be used as a pivot to help future research progress in specific directions, whether it is from a modality perspective or a meta-learning algorithm perspective.\par
\textbf{Granularity of modalities.} The granularity of modalities has not been studied systematically for the FSL/ZSL. Previous studies have already differentiated on the mining of semantic granularity. In some studies, only the simple word embedding or a set of several prescribed attributes~\cite{DBLP:conf/nips/XingROP19,DBLP:conf/wacv/PahdePKN21,hu2018correction} were used to encode the semantics. Nevertheless, in other studies, richer semantics could be employed as well, such as the category labels, richer ‘description level’ semantic sentences or attributes~\cite{DBLP:journals/corr/abs-1906-01905}. The image descriptions with fine-grained text have become a breakthrough point for those studies, which also implies that future research can explore the impact of different modalities, including the granularity's size on the performance of the meta-learning methods.\par
\textbf{Missing modalities.}
As a common data pattern, missing data is more likely to appear in multimodal scenarios in a way that is paired or separated. Considering the inherent task segmentation in the meta-learning framework, more severe missing modalities could also appear in support sets and query sets at the same time. Existing work focuses on the direct application of the meta-learning framework, paying more attention to parameter training or modality fusion, but ignoring the fact that the lack of modalities in multiple scenes could cause unexpected effects. Although several studies~\cite{DBLP:conf/aaai/MaRZTWP21,DBLP:journals/corr/abs-2106-14467} have tried to integrate auxiliary networks to deal with the missing modalities in the whole training, a standardized framework is still to be proposed.\par
\textbf{Multiple modalities.} Current research, whether in terms of methodology or experimental objectives, mostly limits the modalities to visual and textual ones, or dual-modality that appears in pairs. Therefore it is still difficult to apply these methods to multiple modalities other than two. On the one hand, well-aligned multimodal datasets often need to be manually collected and artificially processed to be aligned with multiple modalities. On the other hand, the burden of input parameter fusion or joint training will also increase as the modalities increase. Future research needs to discover ways to construct good benchmark multimodal datasets, instead of only relying on the datasets that are originally designed for image classification. In addition, true multimodal research is supposed to be accompanied by more diverse modality distributions, especially in scenarios such as single-type tasks, cross-tasks, and heterogeneous tasks. It presents challenges to the combination of modalities under different tasks and the meta-learning framework.\par
\textbf{More than data augmentation.}
The literature on learning the data is currently limited, focusing on data augmentation. Future work needs to include more data objectives that can be learned, not only use augmentation methods to learn existing data patterns, but also learn generalized data patterns from similar datasets. Apart from the data objectives, whether some data learning algorithms such as the GAN-based method~\cite{DBLP:journals/corr/abs-1711-04340} can be used for inner learning and optimization in outer learning is still an open challenge for the meta-learning framework.\par
\textbf{More than classification.} Most studies emphasize the use of multimodal data to improve image classification performance, which implements the visual modality as the main modality, accompanied by commonly used auxiliary modalities including text and audio. From a limited scope, this branch of method usually belongs to the category of semantic-based meta-learning, which uses the semantic features of text to enhance the performance of the few-shot classifier by using a convex combination of semantic and visual modalities. The possible future direction is to use the visual modality as an auxiliary modality to perform classification tasks such as text, video, and speech classifications. Such applications put forward high demands on the model for refining image semantics, and need to obtain unique semantics from the visual modality to match the semantic area of other modalities.\par
\textbf{Extend to model-based methods.} Previous taxonomies of meta-learning methods often include the model-based methods as synthesizing models in a feed-forward manner and learning the meta-knowledge of the single model directly. To the best of our knowledge, currently, there is no explicit work to learn such a single model that can be generalized between multimodal tasks. We infer that this may be limited by factors such as the complex process of parameterizing multimodalities and network structures. Future work can try to use the meta-learning models to embed the multimodal dataset into the activation state and predict the test data based on this state.

\bibliographystyle{elsarticle-num-names} 
\bibliography{elsarticle-template-num-names}

\end{document}